\documentclass{article}

\usepackage{microtype}
\usepackage{graphicx}
\usepackage{booktabs} % for professional tables
\usepackage{algorithmic}
\usepackage{multirow}
\usepackage{multicol}
\usepackage{makecell}
\usepackage{amsmath}
\usepackage{graphicx}
\usepackage{caption}
\usepackage{subcaption}
\usepackage{rotating}
\usepackage{enumitem}

\newcommand{\mr}[2]{\multirow{#1}{*}{\begin{tabular}[c]{@{}c@{}}#2\end{tabular}}}
\newcolumntype{C}[1]{>{\centering\let\newline\\\arraybackslash\hspace{0pt}}m{#1}}

\usepackage{amsmath,amsfonts,bm}

\def\eqref#1{equation~\ref{#1}}
\def\1{\bm{1}}

\def\vb{{\bm{b}}}

\def\vw{{\bm{w}}}
\def\vx{{\bm{x}}}
\def\vy{{\bm{y}}}

\def\mK{{\bm{K}}}

\def\mP{{\bm{P}}}

\def\mW{{\bm{W}}}
\def\mX{{\bm{X}}}

\DeclareMathAlphabet{\mathsfit}{\encodingdefault}{\sfdefault}{m}{sl}
\SetMathAlphabet{\mathsfit}{bold}{\encodingdefault}{\sfdefault}{bx}{n}

\newcommand{\E}{\mathbb{E}}

\newcommand{\R}{\mathbb{R}}

\usepackage{hyperref}

\usepackage[accepted]{mlsys2020}

\mlsystitlerunning{Modulating Regularization Frequency for Efficient Compression-Aware Model Training}

\begin{document}

\twocolumn[
\mlsystitle{Modulating Regularization Frequency\\ for Efficient Compression-Aware Model Training}

% unless you've provided the [accepted] option to the mlsys2020


\mlsyssetsymbol{equal}{*}

\begin{mlsysauthorlist}
\mlsysauthor{Dongsoo Lee}{equal,samsung}
\mlsysauthor{Se Jung Kwon}{equal,samsung}
\mlsysauthor{Byeongwook Kim}{equal,samsung}
\mlsysauthor{Jeongin Yun}{samsung}
\mlsysauthor{Baeseong Park}{samsung}
\mlsysauthor{Yongkweon Jeon}{samsung}

\end{mlsysauthorlist}

\mlsysaffiliation{samsung}{Samsung Research, Seoul, Republic of Korea}
\mlsyscorrespondingauthor{Dongsoo Lee}{dongsoo3.lee@samsung.com}

\mlsyskeywords{Machine Learning, MLSys}

\vskip 0.3in

\begin{abstract}
While model compression is increasingly important because of large neural network size, compression-aware training is challenging as it needs sophisticated model modifications and longer training time.
In this paper, we introduce regularization frequency (i.e., how often compression is performed during training) as a new regularization technique for a practical and efficient compression-aware training method.
For various regularization techniques, such as weight decay and dropout, optimizing the regularization strength is crucial to improve generalization in Deep Neural Networks (DNNs).
%For example, when a weight decay factor is far from the optimal value, model accuracy is degraded.
While model compression also demands the right amount of regularization, the regularization strength incurred by model compression has been controlled only by compression ratio.
Throughout various experiments, we show that regularization frequency critically affects the regularization strength of model compression.
Combining regularization frequency and compression ratio, the amount of weight updates by model compression per mini-batch can be optimized to achieve the best model accuracy.
Modulating regularization frequency is implemented by occasional model compression while conventional compression-aware training is usually performed for every mini-batch.
\end{abstract}
]



\printAffiliationsAndNotice{\mlsysEqualContribution} % otherwise use the standard text.

\section{Introduction}

%For Deep Neural Networks (DNNs), a common training method updates weights by gradient descent and explicit weight manipulation through regularization.
Weight regularization is a process adding information to the model to avoid overfitting \cite{deeplearningbook, l2regularization}.
In this paper, we explore weight compression as a form of weight regularization as it severely restricts the search space of weights (i.e., regularized by compression forms).
Moreover, model compression shrinks the effective model size, which is an important regularization principle \cite{deeplearningbook} (note that improved model accuracy by model compression is reported \cite{lotteryscale}).
Weights are regularized in numerous ways by model compression.
For example, each weight can be pruned (e.g., \cite{SHan_2015}) or quantized (e.g., \cite{Greedy_Quan}) to yield a sparse model representation or to reduce the number of bits to represent each weight.

While model compression can be performed without training dataset, compression-aware training can improve model accuracy by reflecting the impact of model compression on the loss function for every mini-batch update \cite{binaryconnect, suyog_prune, ternary2017, Greedy_Quan}.
For such a method, the regularization strength is mainly determined by compression ratio.
Note that for typical regularization schemes, adjusting the regularization strength, such as dropout rate or weight decay factor, is a crucial process to maximize generalization capability of DNNs.
To improve model accuracy given a target compression ratio, we need an additional way to control the regularization strength.
%In general, it would be necessary to weaken the regularization strength if the goal is to increase compression ratio while maintaining the model accuracy.

In this paper, we introduce regularization frequency to represent how many mini-batches are trained without regularization.
Then for weight decay and weight noise insertion, we show that decreasing regularization frequency allows higher weight decay factors or larger amounts of noise per regularization step.
In other words, the overall regularization strength is affected by regularization frequency as well as weight decay factors (or the amount of weight noise).
As a result, a similar amount of average weight updates (determined by both regularization frequency and weight decay factors) is associated with a similar regularization strength, and hence, similar model accuracy.
We demonstrate that the same principle holds for model compression; regularization strength is affected not only by compression ratio but also by regularization frequency.
We verify that our simple model compression techniques (without modifying the underlying training procedures) based on occasional weight regularization can achieve higher compression ratio and higher model accuracy compared to previous techniques that demand substantial modifications to the training process.

Our proposed compression-aware training algorithm enables the followings:
\begin{itemize}%[noitemsep,topsep=0pt,parsep=1pt,partopsep=0pt]
    \item Our compression-aware training technique does not require any modifications to the original training algorithms except an additional regularization method in that weights are occasionally transformed by compression forms.
    \item Computational overhead by compression is not noticeable because such new additional regularization is performed infrequently. Hence, complex compression algorithms are allowed without concerns on training time increase.
    \item We propose an additional regularization hyper-parameter, regularization frequency, to provide larger parameter search space.
    \item Our proposed training method can be a platform to support various kinds of compression techniques including even futuristic ones. Model compression designers can focus on developing new compression architectures without concerns on particular associated training algorithm design.
\end{itemize}

\section{Batch Size Selection}

Since a unit of measurement of regularization frequency is highly correlated with batch size, let us discuss batch size considered for our work.
To overcome some practical issues of gradient descent on non-convex optimization problems, there have been several enhancements such as learning rate scheduling and adaptive update schemes using momentum and update history \cite{overview_optimization}.
Optimizing batch size is another way to yield efficient gradient descent.
Note that large batch size has the advantage of enhancing parallelism of the training system in order to speed up training, critical for DNN research \cite{large_scale_distributed}.
Despite such advantages, small batch size is preferred because it improves generalization associated with flat minima search \cite{largebatch} and other hyper-parameter explorations are more convenient \cite{small_batch}.
Small batch size also affects weight regularization if weight updates for gradient descent and weight regularization are supposed to happen for every mini-batch.
For example, for weight decay conducted for every mini-batch, if batch size is modified, then the weight decay factor should also be adjusted accordingly \cite{decoupledweightdecay}.
In this paper, we assume a reasonably small batch size.

\section{Weight Updates by Model Compression}

\begin{figure}[t]
\centering
	%\begin{minipage}[t]{.40\textwidth}
	\begin{subfigure}[t]{.45\textwidth}
		\includegraphics[width=1\textwidth]{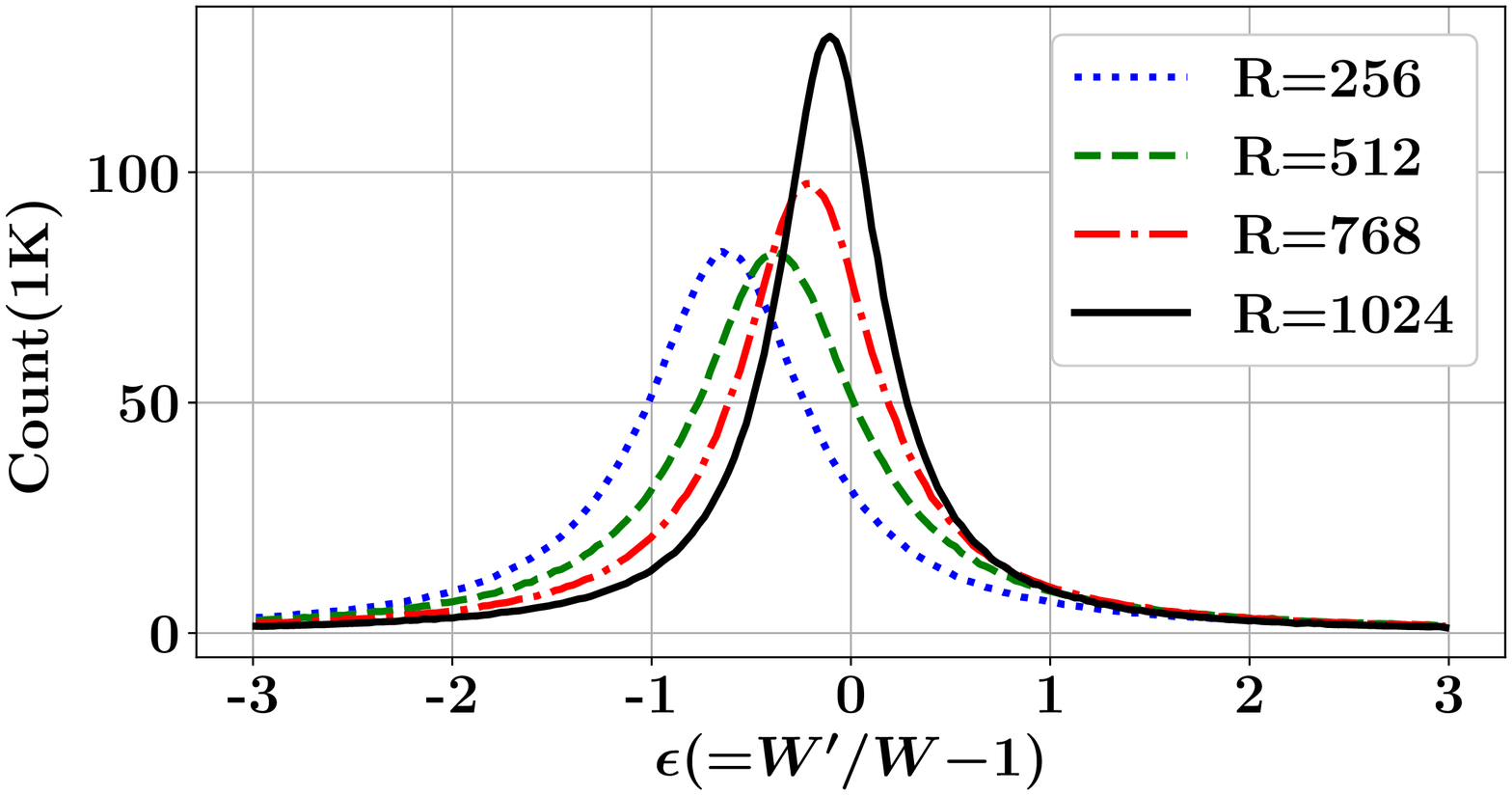}
		\caption{Low-rank approximation via SVD}
	%\end{minipage}
	\end{subfigure}
	\begin{subfigure}[t]{.45\textwidth}
	%\begin{minipage}[t]{.40\textwidth}
		\includegraphics[width=1\textwidth]{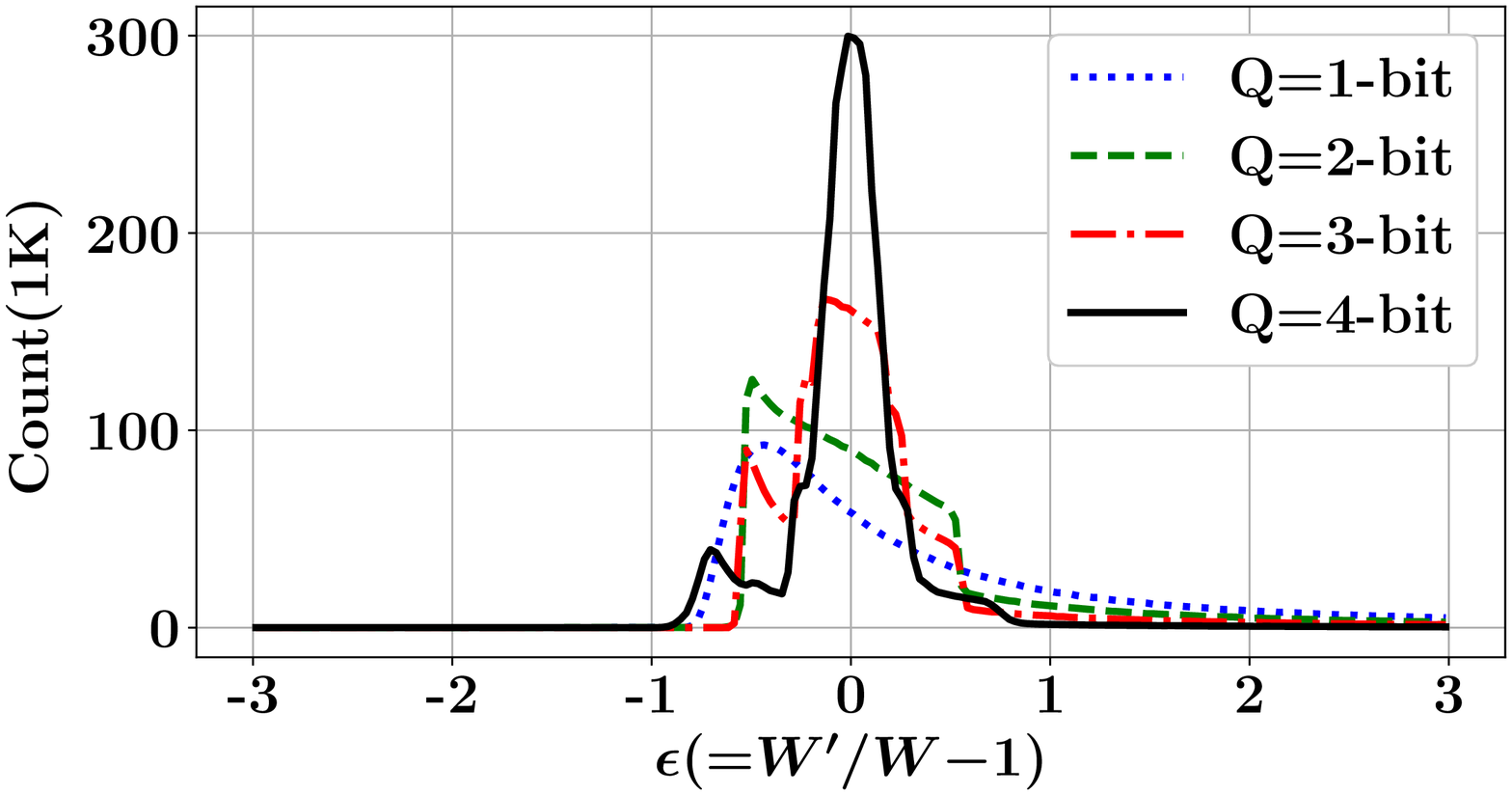}
		\caption{Quantization}
	%\end{minipage}
	\end{subfigure}
	\caption{Distribution of weight noise $\epsilon$ when $R$ is the rank and $Q$ is the number of quantization bits.}
	\label{fig:weight_compression_noise}
\end{figure}

Before investigating the effects of regularization frequency, we first study the relationship between model compression ratio and the weight regularization strength using quantization and singular-value decomposition (SVD) as model compression techniques.
We assume a popular quantization method based on binary codes for which a weight vector $\vw$ is approximated to be $\sum_{i=1}^{q} \alpha_i \vb_i$ for $q$-bit quantization, where $\alpha$ is a scaling factor and $\vb (=\{-1,+1\}^n)$ is a binary vector, and $n$ is the vector size.
The quantization error $|| \vw - \sum_{i}\alpha_i \vb_i||^2$ is minimized by a method proposed by \cite{xu2018alternating} to compute $\alpha$ and $\vb$.
For SVD, a weight matrix $\mW \in\R^{m\times n}$ is approximated to be $\mW' \in\R^{m\times n}$ by minimizing $||\mW - \mW'||$ subject to rank$(\mW') \le R$, where $R$ is the target rank.

For our experiments, we use a synthetic $(2048\times2048$) weight matrix where each element is randomly generated from the Gaussian distribution $\mathcal{N}(\mu\!=\!0, \sigma^2 \!= \!1)$.
Then, we are interested in the amount of change of each weight after quantization and SVD.
Assuming that weight noise through compression is expressed as $\epsilon$ in the form of $w' \!= \!w (1+\epsilon)$, Figure~\ref{fig:weight_compression_noise} shows the distribution of $\epsilon$ with various quantization bits or target ranks.
From $\epsilon$ distributions skewed to be negative, it is clear that weights tend to decay more with higher compression ratio,  along with a wider range of random noise.
Reasonable explanations of Figure~\ref{fig:weight_compression_noise} would include: 1) weights generated from the Gaussian distribution are uncorrelated such that an approximation  step (by compression) using multiple weights would result in noise for each weight, 2) in the case of SVD, elements associated with small eigenvalues are eliminated, 3) averaging effects in quantization reduce the magnitude of large weights.
For weight pruning, $\epsilon$ becomes $-1$ or $0$ (i.e., weight decay for selected weights).
Correspondingly, we study weight decay and weight noise insertion in the next two sections as an effort to gain a part of basic knowledge on improved training for model compression, even though actual model compression would demand much more complicated weight noise models.

\begin{figure*}[t]
\centering
	\includegraphics[width=0.5\linewidth]{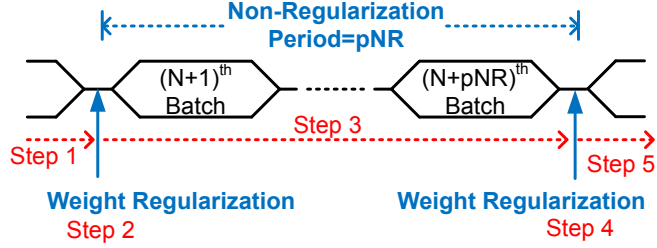}
	\caption{Our proposed modulating regularization frequency scheme when NR period is given as a multiple of batches. 
	%Depending on the loss surface and/or strength of regularization, regularization would lead to step 2 (escaping from a local minimum) or step 5 (returning to a local minimum).
	}
	\label{fig:loss_surface}
\end{figure*}

%\section{Non-Regularization Period}
\section{Non-Regularization Period Study on Weight Decay and Weight Noise Insertion}
%\section{Non-Regularization Period on Weight Decay and Weight Noise Insertion}
%\section{Non-Regularization Period Study on Weight Decay and Noise Insertion}
%\section{Non-Regularization Period Study on Weight Decay/Weight Noise Insertion}

Since weight regularization cannot precede updates for gradient descent, in order to control the frequency of weight regularization, an available option is to skip a few batches without regularization.
In this paper, we propose a new hyper-parameter, called ``Non-Regularization period'' or NR period, to enable occasional regularization and to define the interval of two consecutive regularization events as shown in Figure~\ref{fig:loss_surface}.
NR period is an integer number and expressed as a multiple of batches (from now on, thus, we use `NR period' or $pN\!R$ to represent regularization frequency).

%In the next sections, we focus on the relationship between $pN\!R$ and the strength of weight regularization.

Weight decay is one of the most well-known regularization techniques \cite{three_mecha} and different from $L_2$ regularization in a sense that weight decay is separated from the loss function calculation \cite{decoupledweightdecay}.
Weight decay is performed as
\begin{equation}
\label{eq:weight_decay}
    \vw_{t+1} = (1 - \gamma \theta \vw_t) - \gamma \nabla_{\vw_t} \mathcal{L}(\vw),
\end{equation}
where $\theta$ is a constant weight decay factor.
Weight noise insertion is another regularization technique aiming at reaching flat minima \cite{deeplearningbook, flat_minima}.
For our experiments, we assume that random Gaussian noise is added to weights such that $\vw' \!= \!\vw \!+ \!\bm{\epsilon}$ when $\bm{\epsilon} \!\sim\! \mathcal{N}(0,\eta I)$.

\begin{figure}[t]
\begin{center}
	\begin{minipage}[t]{.40\textwidth}
	    \centering
		\includegraphics[width=1\linewidth]{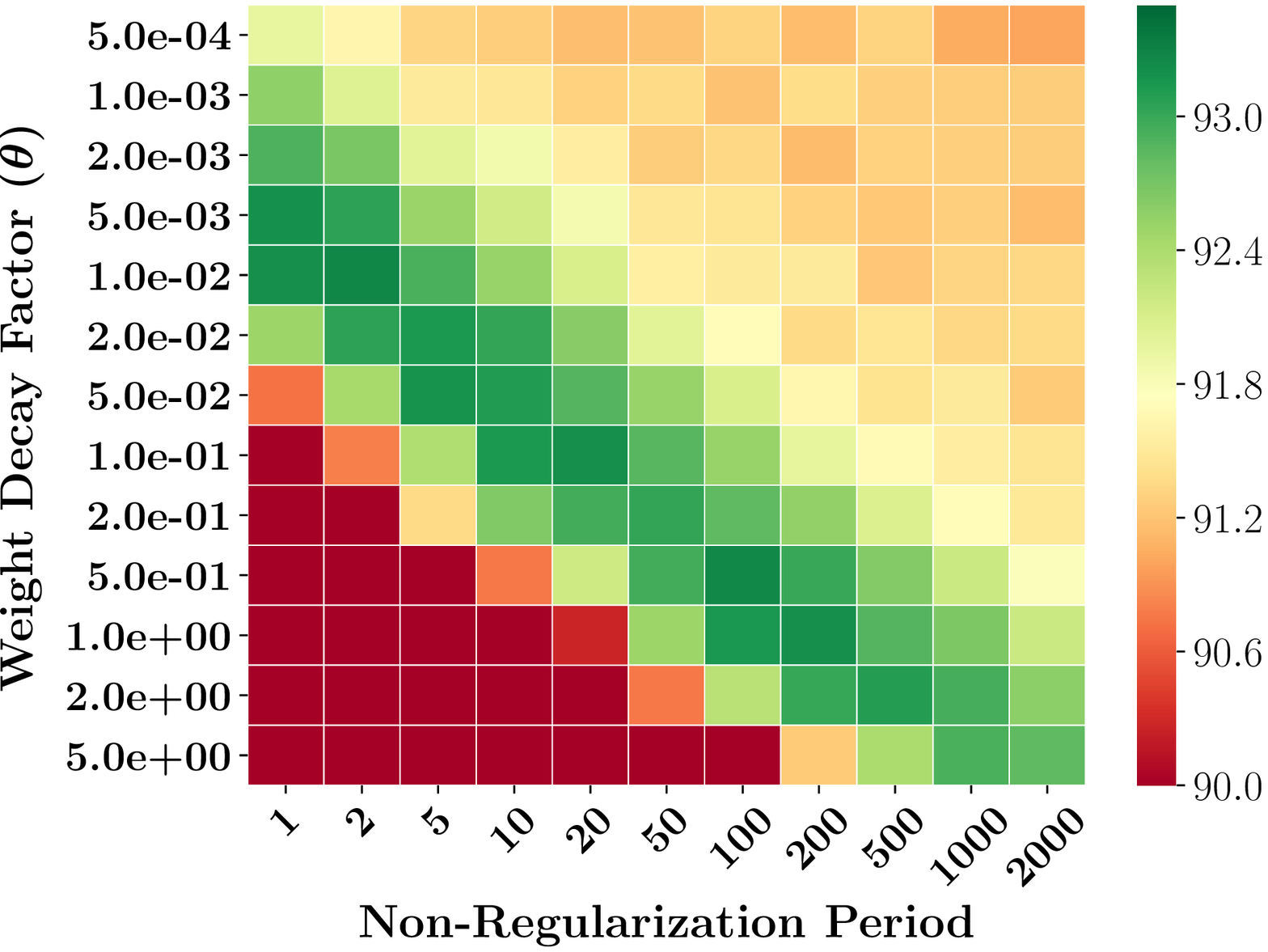}
	\end{minipage}
	\begin{minipage}[t]{.40\textwidth}
	    \centering
		\includegraphics[width=1\linewidth]{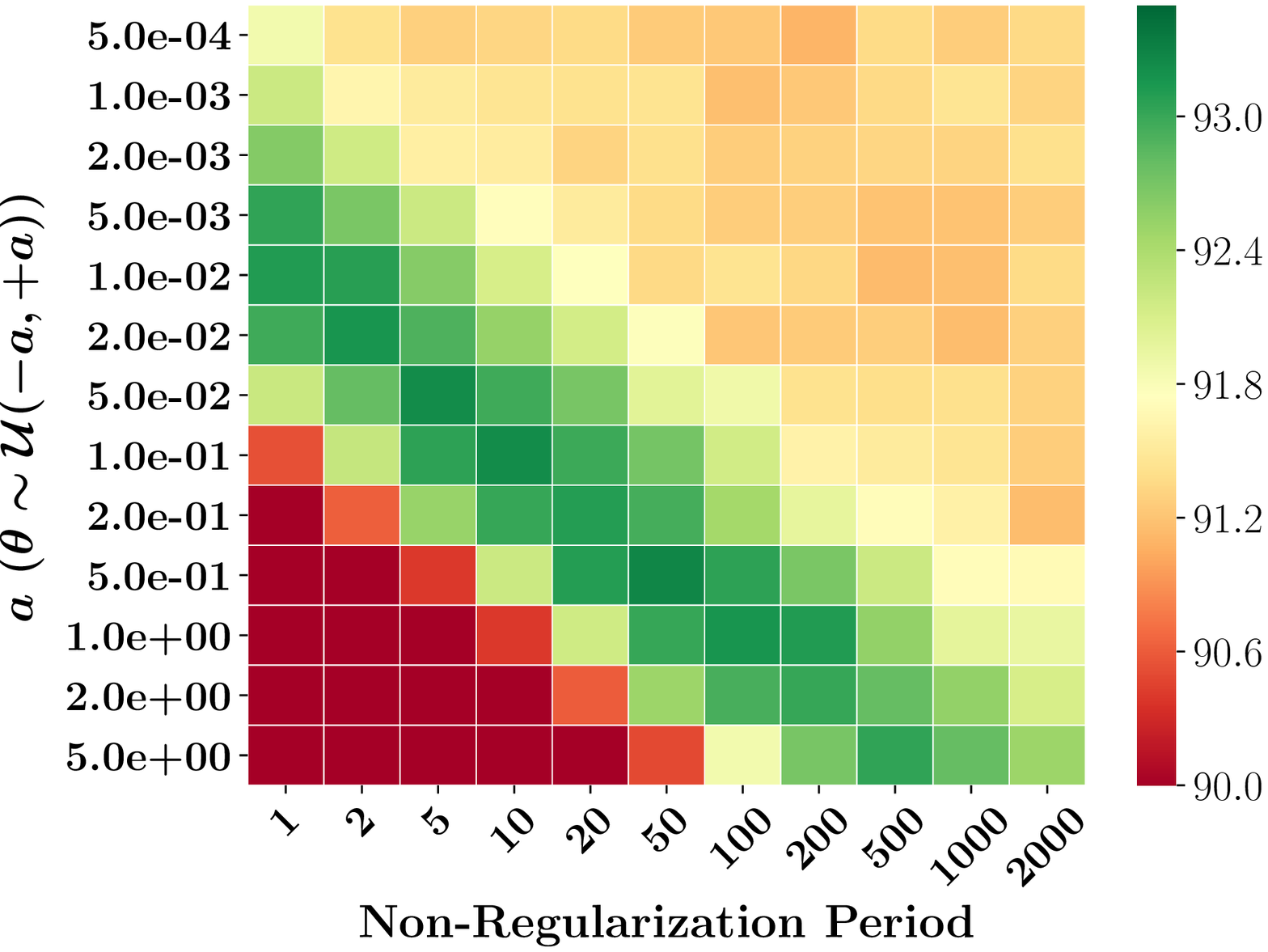}
	\end{minipage}
	\caption{Model accuracy of ResNet-32 on CIFAR-10 using various NR period and amount of weight decay or noise for regularization (original model accuracy without regularization is 92.6\%). (Left): Weight decay. (Right): Uniform weight noise insertion.}
	\label{fig:resnet32_noise}
\end{center}
\end{figure}

\begin{figure}[t]
\begin{center}
	\begin{minipage}[t]{.40\textwidth}
	    \centering
		\includegraphics[width=1\linewidth]{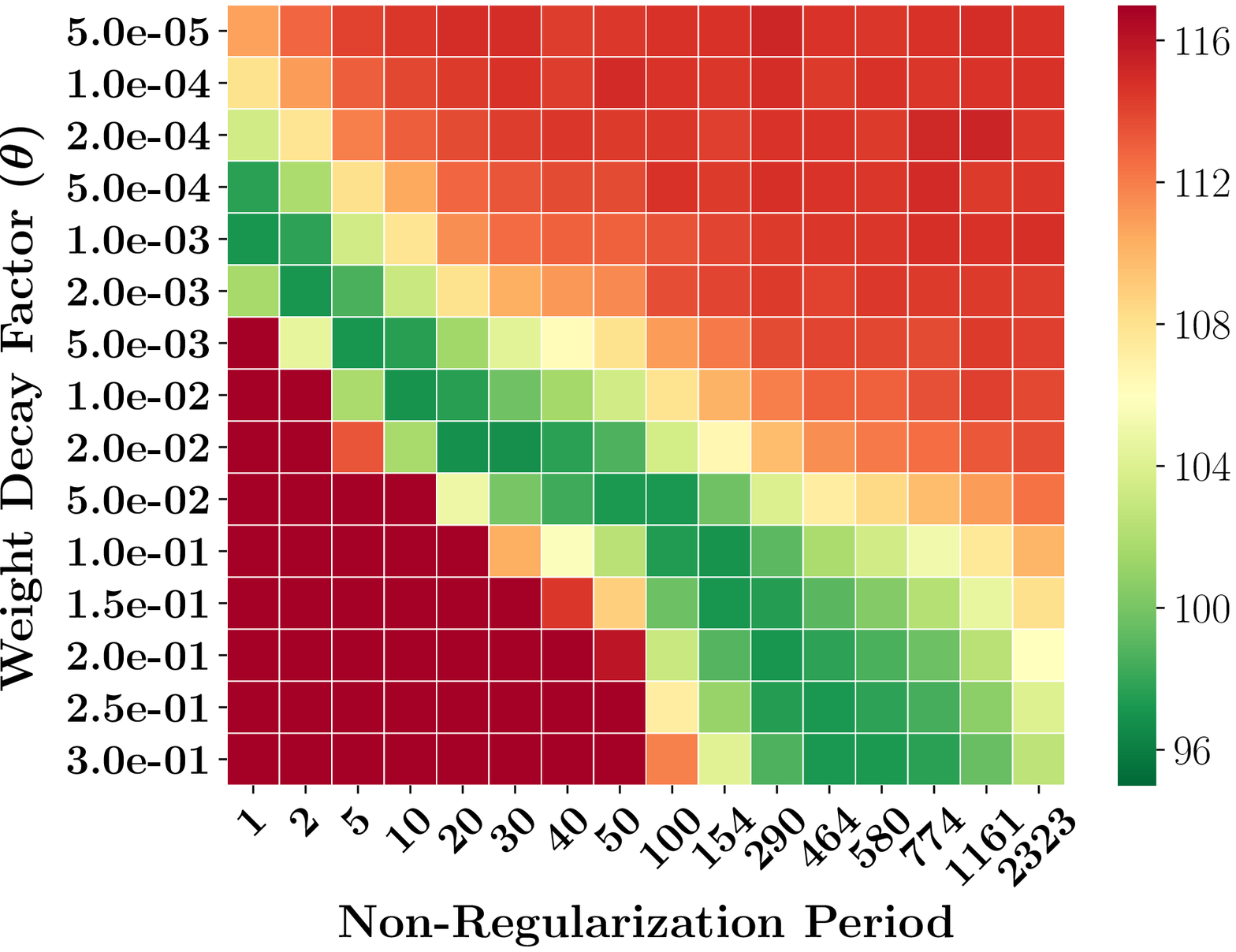}
	\end{minipage}
	\begin{minipage}[t]{.40\textwidth}
	    \centering
		\includegraphics[width=1\linewidth]{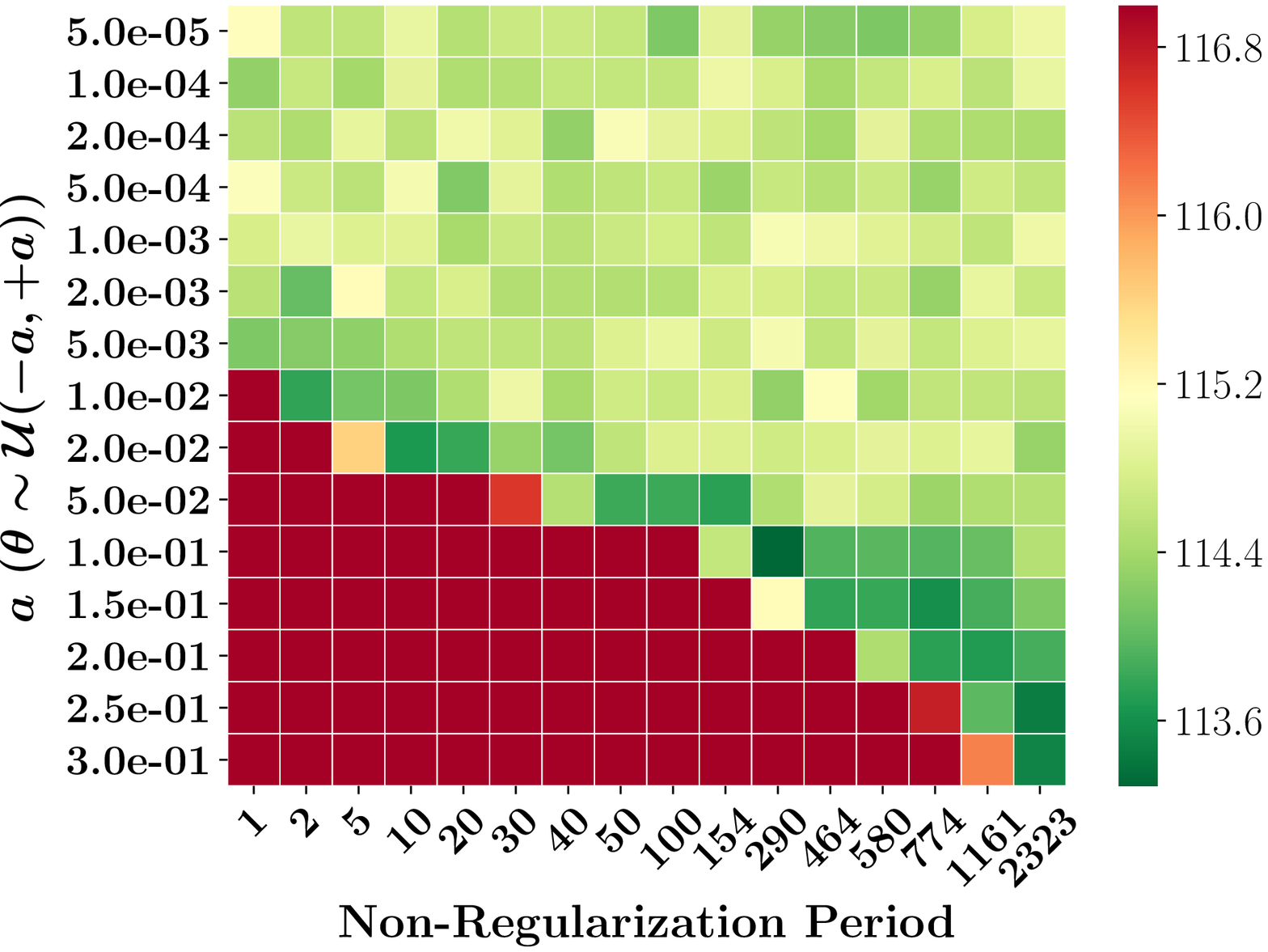}
	\end{minipage}
	\caption{Perplexity of LSTM model on PTB dataset using various NR period and amounts of weight decay or noise. (Left): Weight decay. (Right): Uniform weight noise insertion.}
	\label{fig:ptb_noise}
\end{center}
\end{figure}

We study the impact of NR period on weight decay and weight noise insertion using ResNet-32 on CIFAR-10 model \cite{resnet} and a long short-term memory (LSTM) model on PTB dataset \cite{PTB_google}.
For the LSTM model, we use 2 layers with 200 hidden units and the hyper-parameter set introduced by \cite{PTB_google}.
For the weight noise model, we plug $\theta \sim \mathcal{U}(-a, +a)$ (uniform distribution) into Eq.~(\ref{eq:weight_decay}) to simplify the experiments.
Figure~\ref{fig:resnet32_noise} shows model accuracy of ResNet-32 given different NR period and weight decay factors.
For both weight decay and weight noise insertion, the choice of $\theta$ (representing the amount of weight regularization for every $pN\!R$) has a clear correlation with NR period (refer to Appendix for training and test accuracy graphs).
If we wish to apply larger $\theta$, then weight regularization should be conducted less frequently (i.e., larger weight decay factor requires longer NR period) to optimize the regularization effect and achieve high model accuracy.
For similar model accuracy, weight decay factor can be approximately 1,000 times larger with $pN\!R \approx 1,000$ in Figure~\ref{fig:resnet32_noise} and Figure~\ref{fig:ptb_noise}.
Similar observations are discovered by the LSTM model on PTB as shown in Figure~\ref{fig:ptb_noise}.
Lower perplexity (indicating better generalization) is obtained when the NR period increases as $\theta$ becomes larger for each regularization event.
For weight decay, increasing $\theta$ by longer $pN\!R$ may not be significant because of similar model accuracy.
On the other hand, \textbf{increasing model compression ratio by longer $pN\!R$ should be significant as we show in the next section.}
%Note that compared with a conventional weight decay factor selection (i.e., $\theta$ in Eq.~(\ref{eq:weight_decay}) when $pN\!R=1$), for similar model accuracy, weight decay factor can be approximately 1,000 times larger with $pN\!R \approx 1,000$ in Figure~\ref{fig:resnet32_noise} and Figure~\ref{fig:ptb_noise}.
%Let $e_w$ be the mean absolute difference given as ($=\E [|\vw - \vw' |]$), where $\vw'$ is a weight vector after regularization.
%Figure~\ref{fig:resnet32_noise} and Figure~\ref{fig:ptb_noise} imply that $e_w / pN\!R$ (or $\theta / pN\!R$) determines the regularization strength.
%Consequently, optimal $e_w$ depends on $pN\!R$, and hence, we argue that a wide exploration of various NR period values and regularization hyper-parameters (such as $\theta$) is necessary in order to best utilize regularization effects on a DNN model while previous attempts employ $pN\!R=1$ only.
%For model compression, high compression ratio corresponds to large $\theta$ of weight decay, and thus, increasing compression ratio requires large $pN\!R$ value that is indeed demonstrated thoroughly in the next section.
%In addition, \textit{the observation that stronger weight regularization is enabled by longer $pN\!R$ is our basic training principle for model compression.}

\section{NR Period for Model Compression}
    
As discussed, weight compression incurs a much more complicated weight regularization model than weight decay or uniform weight noise insertion because 1) as shown in Figure~\ref{fig:weight_compression_noise}, diversified noise models need to be combined to describe weight regularization after model compression and 2) compression-aware training methods would reduce the strength of weight regularization as training is performed with more epochs and weights converge to a compressed form.
Nonetheless, we can conjecture that the best training scheme for model compression may require the condition of $pN\!R \neq 1$ that can be empirically justified.
    
\begin{figure*}[t]
    \centering
    \begin{subfigure}{0.98\textwidth}
        \centering
	    \includegraphics[width=0.92\linewidth]{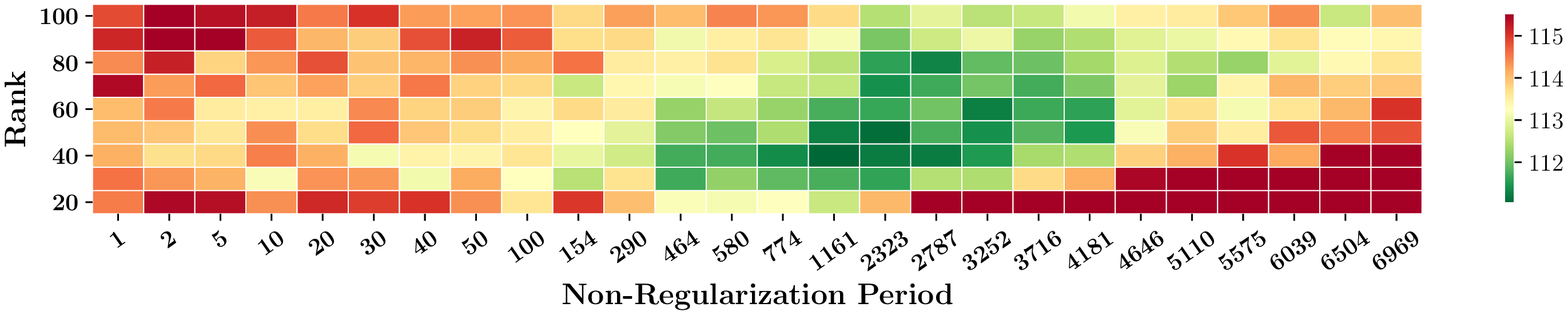}
	    \caption{Perplexity when the weights are compressed by SVD.}
	    \label{fig:ptb_svd}
	\end{subfigure}
	\centering
	\begin{subfigure}{0.98\textwidth}
        \centering
	    \includegraphics[width=0.92\linewidth]{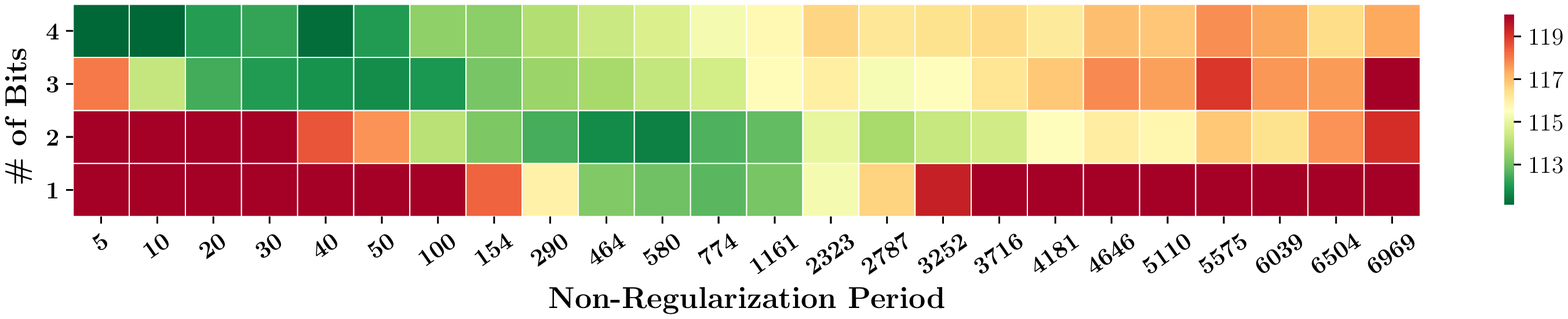}
	    \caption{Perplexity when the weights are compressed by quantization based on binary codes.}
	    \label{fig:ptb_quant}
	\end{subfigure}
	\centering
	\begin{subfigure}{0.98\textwidth}
        \centering
        \includegraphics[width=0.92\linewidth]{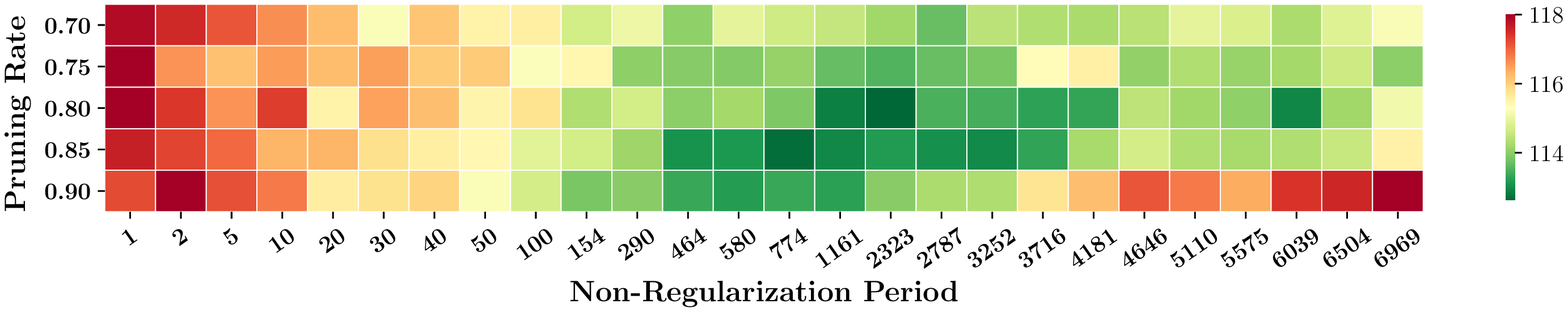}
	    \caption{Perplexity when the weights are compressed by magnitude-based pruning.}
	    \label{fig:app_ptb_pruning}
    \end{subfigure}
	\caption{Model accuracy of an LSTM model on PTB compressed by quantization, low-rank approximation or pruning. Original perplexity without model compression is 114.6. For more details, refer to Appendix.}
	\label{fig:ptb_pnr_weight_relationship}
\end{figure*}

\begin{figure*}[t]
    \centering
    \begin{subfigure}{0.32\textwidth}
        \centering
	    \includegraphics[width=1.0\textwidth]{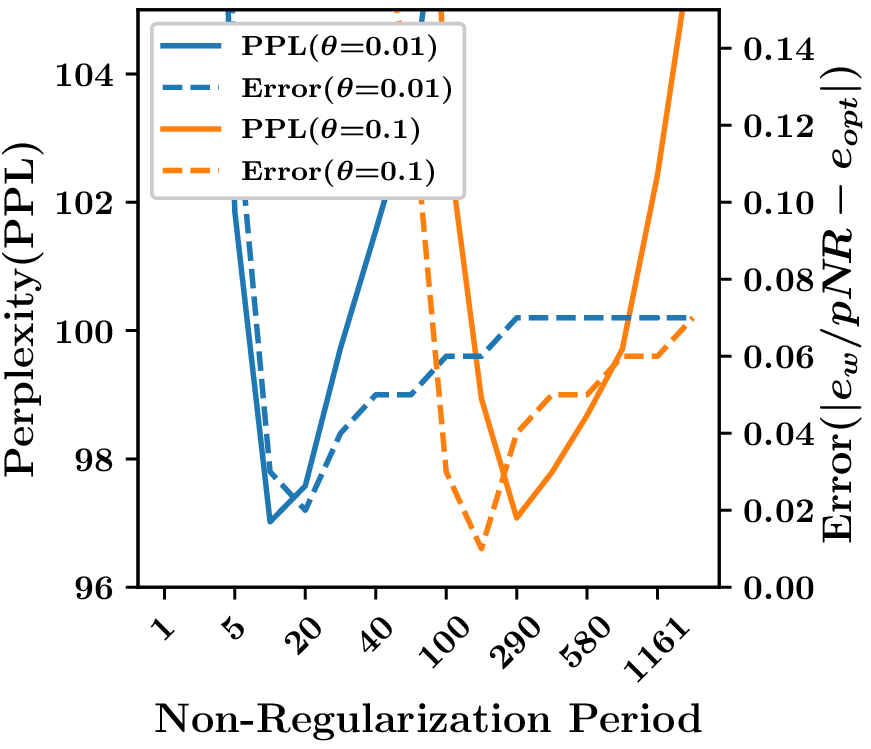}
	    \caption{Weight Decay ($e_{opt}{=}0.0007$)}
	\end{subfigure}
	\begin{subfigure}{0.32\textwidth}
        \centering
	    \includegraphics[width=1.0\textwidth]{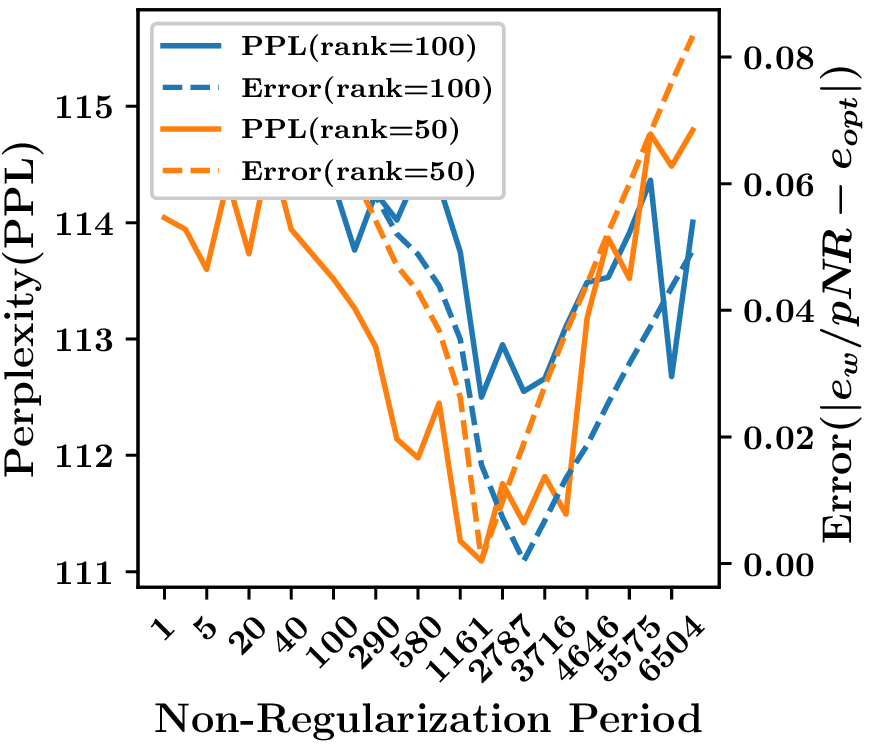}
	    \caption{SVD ($e_{opt}{=}0.08$)}
	\end{subfigure}
	\begin{subfigure}{0.32\textwidth}
        \centering
        \includegraphics[width=1.0\textwidth]{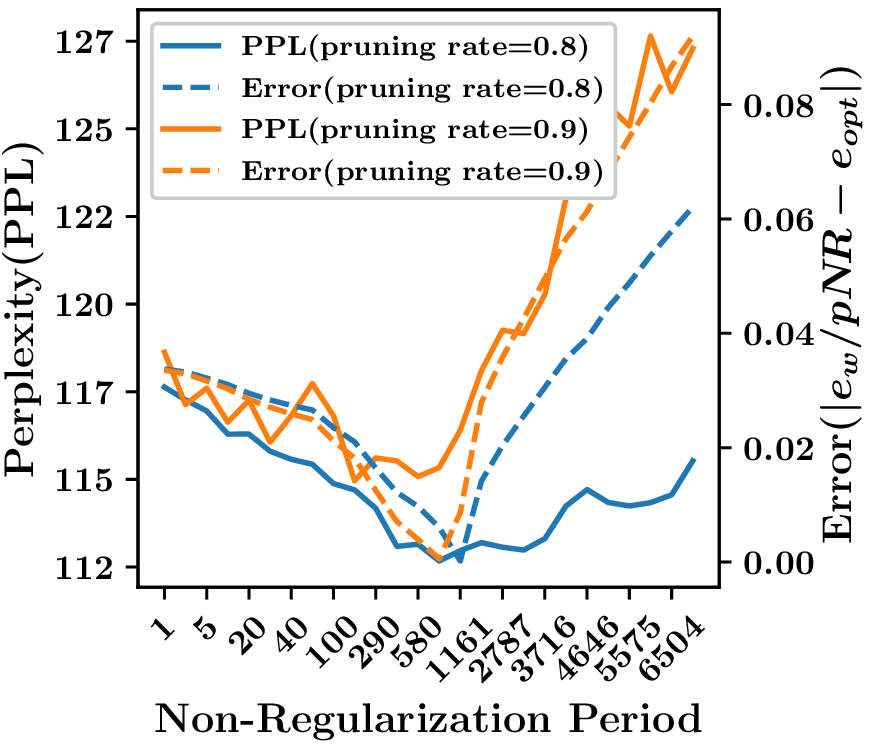}
	    \caption{Pruning ($e_{opt}{=}0.035$)}
    \end{subfigure}
	\caption{Model accuracy and regularization error (defined as the difference between $e_w/pN\!R$ and $e_{opt}$) using PTB LSTM model when weights are regularized by weight decay, SVD, or pruning.}
	\label{fig:ptb_pnr_err}
\end{figure*}

\begin{figure*}[t]
    \centering
    \begin{subfigure}{0.32\textwidth}
        \centering
	    \includegraphics[width=1.0\textwidth]{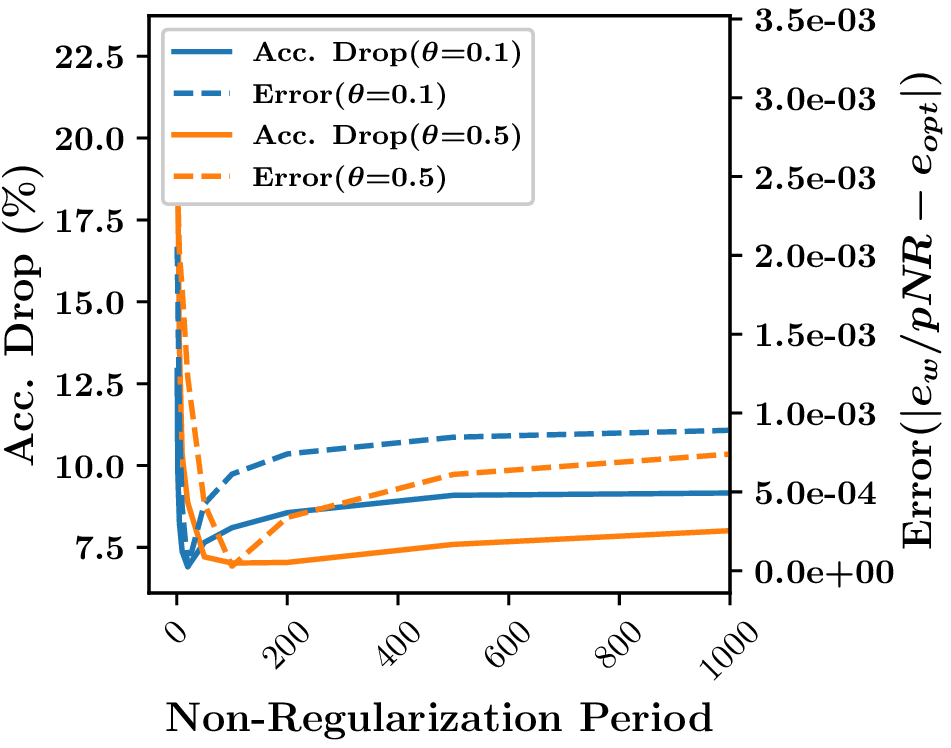}
	    %\caption{Weight Decay ($e_{opt}{=}0.000009455$)}
	    \caption{Weight Decay ($e_{opt}{=}$9.46e-06)}
	\end{subfigure}
	\begin{subfigure}{0.32\textwidth}
        \centering
	    \includegraphics[width=1.0\textwidth]{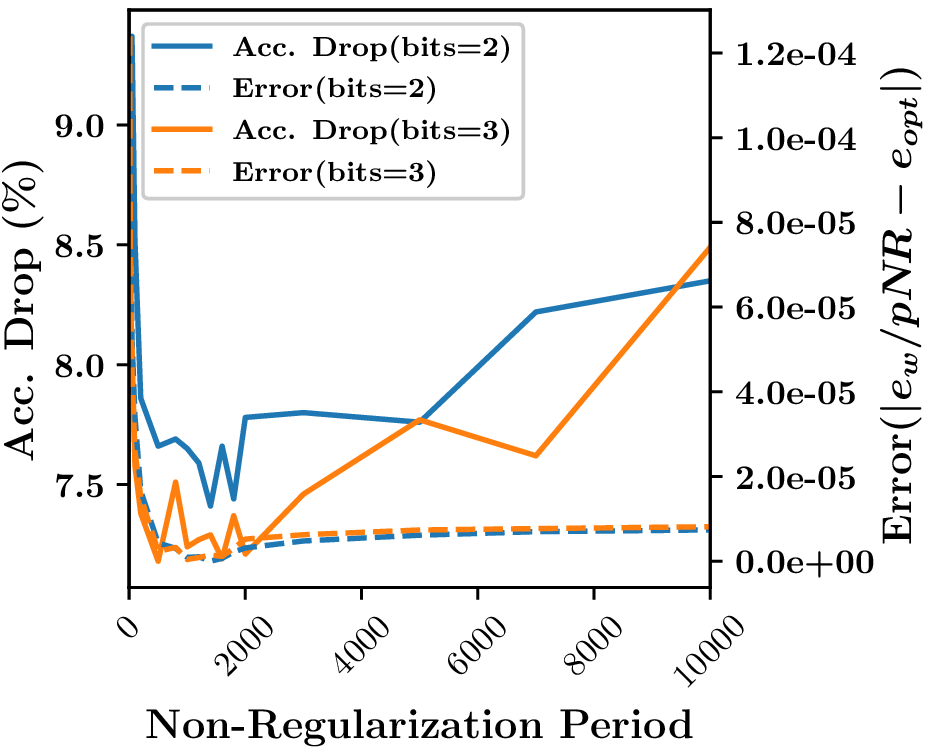}
	    \caption{Quantization ($e_{opt}{=}$9.38e-06)}
	\end{subfigure}
	\begin{subfigure}{0.32\textwidth}
        \centering
        \includegraphics[width=1.0\textwidth]{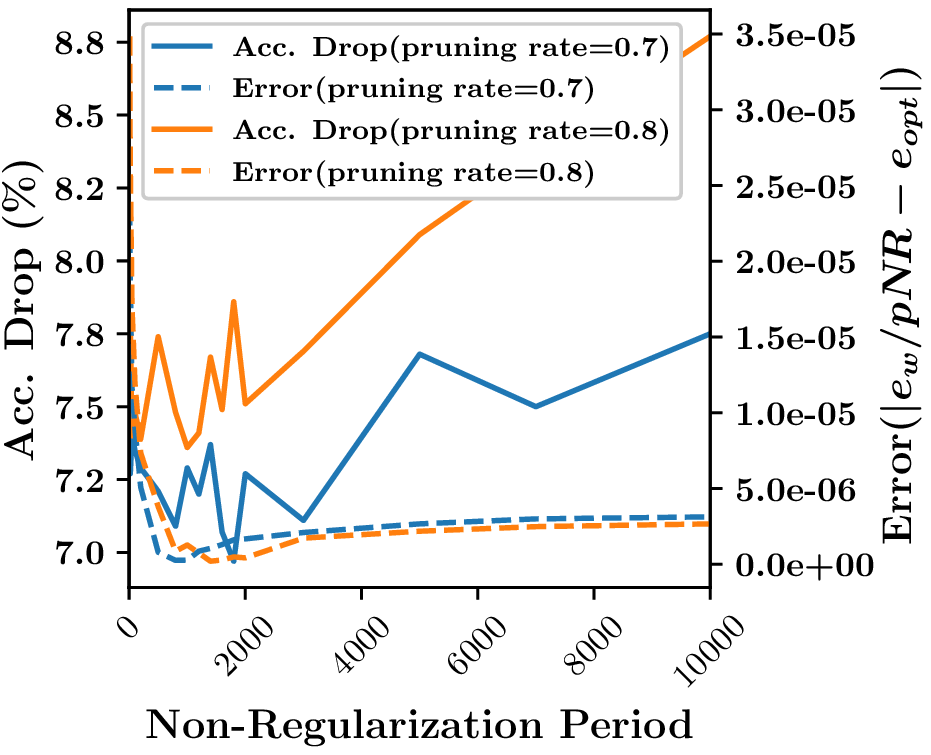}
	    \caption{Pruning ($e_{opt}{=}$4.25e-6)}
    \end{subfigure}
	\caption{Test accuracy drop and regularization error using ResNet-32 on CIFAR-10 when weights are regularized by weight decay, quantization, or pruning.}
	\label{fig:resnet_pnr_err}
\end{figure*}

We apply weight quantization, low-rank approximation (SVD), and pruning to an LSTM model on PTB that we selected for the previous section.
We do not modify underlying training principles and use the following simple strategy:

\framebox{\parbox{0.97\linewidth}{
\begin{enumerate}[noitemsep,topsep=0pt,parsep=1pt,partopsep=0pt]
    \item [(1)] Train the model for $pN\!R$ batches \\(as if model compression is not being considered.)
    \item [(2)] Perform weight compression in the form of \\$\vw' = h(\vw )$.
    \item [(3)] With new full-precision weight $\vw'$, \\repeat the above two steps.
\end{enumerate}
}}

$h(\vw)$ can be a magnitude-based pruning (i.e., $h(w)$=$w$ if $|w|$ is larger than a certain threshold, or $h(w)$=$0$, otherwise), $\alpha \vb$ for quantization, SVD function, or even as-yet undiscovered functions.

Figure~\ref{fig:ptb_pnr_weight_relationship} shows model accuracy associated with a number of different sets of $pN\!R$ and model compression strength (i.e., target rank for low-rank approximation and the number of quantization bits).
Notice that the optimal $pN\!R$ for the best model accuracy is definitely larger than $1$.
To explain how Figure~\ref{fig:ptb_pnr_weight_relationship} is aligned with the previous section, we investigate the relationship between model accuracy and the average of $e_w/pN\!R$ ($e_w = \E [| \vw - \vw' |]$, where $\vw'$ is a weight vector after weight decay, SVD, or pruning) throughout the entire training.
We first optimize $e_w/pN\!R$ $(=e_{opt})$ to achieve the best model accuracy.
Following Figure~\ref{fig:resnet32_noise} and \ref{fig:ptb_noise}, let us assume $e_{opt}$ to be constant regardless of compression ratio or decay factor, and obtained by finding hyper-parameter sets associated with maximum model accuracy in Figure~\ref{fig:ptb_noise} and Figure~\ref{fig:ptb_pnr_weight_relationship} and by taking the average of corresponding $e_w$ values.
When regularization error is defined to be $| e_w/pN\!R - e_{opt}|$, Figure~\ref{fig:ptb_pnr_err} shows test perplexity and regularization error of PTB LSTM model with different $pN\!R$.
Such defined regularization error is affected by $pN\!R$ as shown in Figure~\ref{fig:ptb_pnr_err}, and indeed, when regularization error approaches to the minimum (i.e., zero) we gain improved model accuracy.
Unlike weight decay where $e_w$ is directly computed by decay factors, for model compression techniques, $e_w$ is not directly related to compression-related hyper-parameters (such as ranks and pruning rates).
As a result, while Figure~\ref{fig:ptb_noise} shows a clear correlation between decay factors and $pN\!R$ for best model accuracy, Figure~\ref{fig:ptb_pnr_weight_relationship} suggests that compression ratio and $pN\!R$ are weakly correlated.
Hence, $pN\!R$ is a hyper-parameter to be determined empirically for model compression.
Nonetheless, the optimal $pN\!R$ is definitely larger than 1, as shown in Figure~\ref{fig:ptb_pnr_err}, and decoupled from batch size selection.
That means weight regularization for model compression needs to be conducted much less frequently compared with gradient descent since batch size selection considers generalization ability of gradient descent, not regularization effects.

For ResNet-32 on CIFAR-10, we also find $e_{opt}$ and investigate the relationship between model accuracy and $pN\!R$ as shown in Figure~\ref{fig:resnet_pnr_err}.
Similar to the case of PTB LSTM model, ResNet-32 presents a particular $pN\!R$ that minimizes regularization error.
It is clear that for both PTB LSTM and ResNet-32, optimal $pN\!R$ is definitely larger than `1' despite some variation on model accuracy.
We summarize our empirical observations as follows:

\begin{itemize}%[noitemsep,topsep=0pt,parsep=1pt,partopsep=0pt]
    \item Unlike conventional wisdom, a wide range of weight decay factors is allowed since we can adjust $pN\!R$ to optimize the regularization strength.
    \item For each weight decay factor selected, there is an optimal $pN\!R$ to maximize model accuracy.
    \item Similarly, for each compression ratio, a particular $pN\!R$ presents the best model accuracy.
    \item Such $pN\!R$ is a hyper-parameter that is empirically searched.
\end{itemize}

From our extensive experiments, optimal $pN\!R$ is usually searched in the range from 10 to 1000 for model compression.
Large $pN\!R$ provides a benefit of less amount computations for model compression.
Especially when the compression method is based on iterative mathematical principles (such as SVD \cite{SVD2013} or quantization \cite{xu2018alternating}), large $pN\!R$ can save training time significantly.

\section{Comparison with Previous Model Compression Techniques}

In this section, we compare some of previous model compression techniques with our compression scheme that introduces $pN\!R$ and obviates special training algorithm modifications.
Due to the space limit, please refer to Appendix for more experimental results with ImageNet and PTB.

\subsection{Fine-Grained Weight Pruning}

The initial attempt of pruning weights was to locate redundant weights by computing the Hessian to calculate the sensitivity of weights to the loss function \cite{optimalbrain}.
However, such a technique has not been considered to be practical due to significant computation overhead for computing the Hessian.
Magnitude-based pruning \cite{SHan_2015} has become popular because one can quickly find redundant weights by simply measuring the magnitude of weights.
Since then, numerous researchers have realized a higher compression ratio largely by introducing Bayesian inference modeling of weights accompanying supplementary hyper-parameters.
For example, dynamic network surgery (DNS) \cite{DNS} permits weight splicing when a separately stored full-precision weight becomes larger than a certain threshold.
Optimizing splicing threshold values, however, necessitates extensive search space exploration, and thus, longer training time.
Variational dropout method \cite{sparseVD} introduces an explicit Bayesian inference model for a prior distribution of weights, which also induces various hyper-parameters and increased computational complexity.

We perform magnitude-based pruning at every $pN\!R$ step.
As a result, even though weights are pruned and replaced with zero at $pN\!R$ steps, pruned weights are still updated in full precision during NR period.
If the amount of updates of a pruned weight grows large enough between two consecutive regularization steps, then the weight pruned at the last $pN\!R$ step may not be pruned at the next $pN\!R$ step.
Such a feature (i.e., pruning decisions are not fixed) is also utilized for weight splicing in DNS \cite{DNS}.
Weight splicing in DNS relies on a hysteresis function (demanding sophisticated fine-tuning process with associated hyper-parameters) to switch pruning decisions.
Pruning decisions through our scheme, on the other hand, are newly determined at every $pN\!R$ step.

We present experimental results with LeNet-5 and LeNet-300-100 models on MNIST dataset which are also reported by \cite{DNS, sparseVD}.
LeNet-5 consists of 2 convolutional layers and 2 fully connected layers while 3 fully connected layers construct  LeNet-300-100.
We train both models for 20000 steps using Adam optimizer where batch size is 50.
All the layers are pruned at the same time and the pruning rate increases gradually \cite{suyog_prune}.
%following the equation introduced in \cite{suyog_prune}:
%\begin{equation}
%p_t = p_f + \left( p_i - p_f \right) \left( 1 - \frac{t-t_i}{t_f - t_i}\right)^{E} ,
%\label{eq:eq1}
%\end{equation}
%where $E$ is a constant, $p_f$ is the target pruning rate, $p_i$ is the initial pruning rate, $t$ is the current step, and the pruning starts at training step $t_i$ and reaches $p_f$ at training step $t_f$.
%After $t_f$ steps, pruning rate is maintained to be $p_f$.
%For LeNet-5 and LeNet-300-100, $t_i$, $p_i$, $E$ are 8000 (step), 25(\%), and 7, respectively.
%$t_f$ is 12000 (step) for LeNet-5 and 13000 (step) for LeNet-300-100.
%Note that these choices are not highly sensitive to test accuracy as discussed in \cite{suyog_prune}.
We exclude dropout to improve the accuracy of LeNet-300-100 and LeNet-5 since pruning already works as a regularizer \cite{SHan_2015, dropconnect}.
We keep the original learning schedule and the total number of training steps (no additional training time for model compression).

\begin{table}[t]
    \small
    \caption{Pruning rate and accuracy comparison using LeNet-300-100 and LeNet-5 models on MNIST dataset. DC (Deep Compression) and Sparse VD represent a magnitude-based technique \cite{deepcompression} and variational dropout method \cite{sparseVD}, respectively.}
    \label{table:lenet_pruning_comparison}
\begin{center}
\setlength\tabcolsep{3.0pt}
    \begin{tabular}{c c c c c c c}
    \hline
    \multirow{2}{*}{Model} & \multirow{2}{*}{Layer} & Weight & \multicolumn{4}{c}{Pruning Rate (\%)}\\
    \cline{4-7}
    & & Size & DC & DNS & SparseVD & Ours \\
    \hline
     & FC1 & 235K & 92 & 98.2 & 98.9 & 98.9 \\
    LeNet & FC2 & 30K & 91 & 98.2 & 97.2 & 96.0 \\
    -300-100 & FC3 & 1K & 74 & 94.5 & 62.0 & 62.0 \\
    \cline{2-7}
    & Total & 266.2K & 92 & 98.2 & 98.6 & 98.4 \\
    \hline
    \multirow{5}{*}{LeNet-5} & Conv1 & 0.5K & 34 & 85.8 & 67 & 60.0 \\
    & Conv2 & 25K & 88 & 96.9 & 98 & 97.0 \\
    & FC1 & 40K & 92 & 99.3 & 99.8 & 99.8 \\
    & FC2 & 5K & 81 & 95.7 & 95 & 95.0 \\
    \cline{2-7}
    & Total & 430K & 92 & 99.1 & 99.6 & 99.5 \\
    \hline
    \\
    \end{tabular}
    
    \begin{tabular}{c c c c c}
        \hline
        \multirow{2}{*}{Model} & \multicolumn{4}{c}{Accuracy (\%)} \\
        \cline{2-5}
        & DC & DNS & Sparse VD & DeepTwist \\
        \hline
        LeNet-300-100 & 98.4 & 98.0 & 98.1 & 98.1 \\
        LeNet-5 & 99.2 & 99.1 & 99.2 & 99.1 \\
        \hline
    \end{tabular}
\end{center}
\end{table}

\begin{table*}[t]
    \caption{Test accuracy (average of 10 runs for each choice of $pN\!R$) of pruned LeNet-5 model. Pruning rates are described in Table \ref{table:lenet_pruning_comparison}.}
    \label{table:lenet_step_exp}
\begin{center}

    \begin{tabular}{c | c c c c c c c c c}
    \hline
    NR Period ($pN\!R$) & 1 & 2 & 5 & 10 & 50 & 100 & 200 & 500 \\
    \hline
    Accuracy (\%) & 99.00 & 99.06 & 99.06 & 99.11 & 99.05 & 98.98 & 98.72 & 96.52 \\
    \hline
    \end{tabular}
\end{center}
\end{table*}

Table~\ref{table:lenet_pruning_comparison} presents the comparison on pruning rates (see Appendix for test accuracy that is almost the same among all selected schemes).
Despite the simplicity, our pruning scheme produces higher pruning rate compared with DNS and similar compared with variational dropout technique which involves much higher computational complexity.
For Table \ref{table:lenet_pruning_comparison}, we use $pN\!R$=10 for LeNet-5 and $pN\!R$=5 for LeNet-300-100.

We investigate how sensitive $pN\!R$ is to the test accuracy when the other parameters (such as pruning rates, learning rate, and total training time) are fixed.
As shown in Table 2, for a wide range of $pN\!R$, the test accuracy has negligible fluctuation\footnote{Even though we cannot show such a sensitivity study for all of the remaining experiments in this paper, $pN\!R$ has also shown low sensitivity to the accuracy even for other models and compression techniques.}.
Too large $pN\!R$ would result in 1) too little weight distortion, 2) coarse-grained gradual pruning, and 3) unnecessarily large updates for correctly pruned weights.
On the other hand, too small $pN\!R$ may yield excessive amounts of weight distortion and reduce the opportunity for the pruned weights to recover.

We apply $pN\!R$-based pruning to an RNN model to verify the effectiveness of $pN\!R$.
We choose an LSTM model \citep{PTB_google} on the PTB dataset \citep{Marcus_1993}.
Following the model structure given in \cite{PTB_google}, our model consists of an embedding layer, 2 LSTM layers, and a softmax layer.
The number of LSTM units in a layer can be 200, 650, or 1500, depending on the model configurations (referred as small, medium, and large model, respectively).
The accuracy is measured by Perplexity Per Word (PPW), denoted simply by perplexity in this paper.
%We apply gradual pruning with $E=3$, $t_i=0$, $p_i=0$, $t_f = 3^{rd}$ epoch (for medium) or $5^{th}$ epoch (for large) to the pre-trained PTB models.
$pN\!R$-based pruning for the PTB models is performed gradually using $pN\!R =100$ and the initial learning rate is 2.0 for the medium model (1.0 for pre-training) and 1.0 for the large model (1.0 for pre-training) while the learning policy remains to be the same as in \cite{PTB_google}.

\begin{table*}[t!]
    \caption{Comparison on perplexity using various pruning rates. $p_f$ is the target pruning rates for the embedded layer, LSTM layer, and softmax layer.}
    \label{table:ptb_pruning}
\begin{center}

    \begin{tabular}{c c c c c c c c c}
    \hline
    \multirow{2}{*}{Model Size} & \multirow{2}{*}{Pruning Method} & & \multicolumn{6}{c}{Perplexity} \\
    \cline{3-9}
     & & $p_f$=& 0\% & 80\% & 85\% & 90\% & 95\% & 97.5\% \\ 
    \hline
    Medium & \citep{suyog_prune} & & 83.37 & 83.87 & 85.17 & 87.86 & 96.30 & 113.6 \\
    (19.8M) & DeepTwist & & 83.78 & 81.54 & 82.62 & 84.64 & 93.39 & 110.4 \\
    \hline
    Large & \citep{suyog_prune} & & 78.45 & 77.52 & 78.31 & 80.24 & 87.83 & 103.20 \\
    (66M) & DeepTwist & & 78.07 & 77.39 & 77.73 & 78.28 & 84.69 & 99.69 \\
    \hline
    \end{tabular}
\end{center}
\end{table*}

\begin{figure}[t]
	\begin{minipage}{.5\textwidth}
	\centering
		\includegraphics[width=0.8\linewidth]{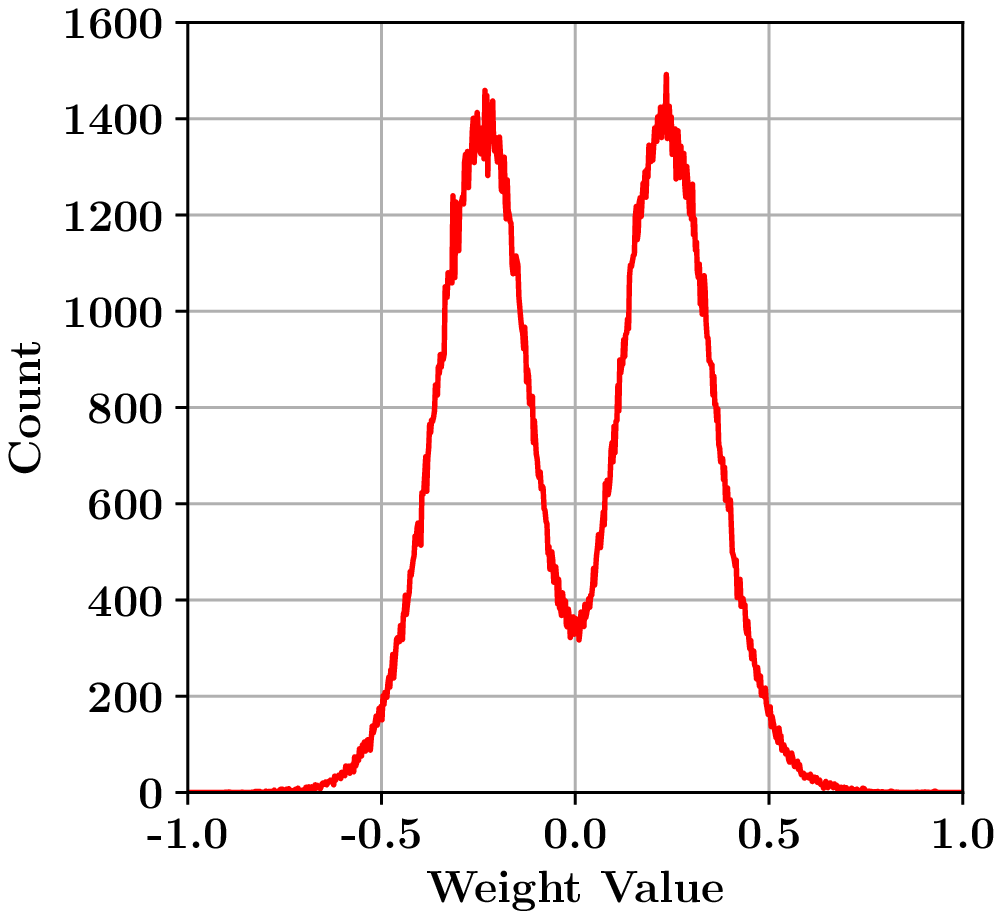}
		
	\end{minipage}
	\begin{minipage}{.5\textwidth}
	\centering
		\includegraphics[width=0.8\linewidth]{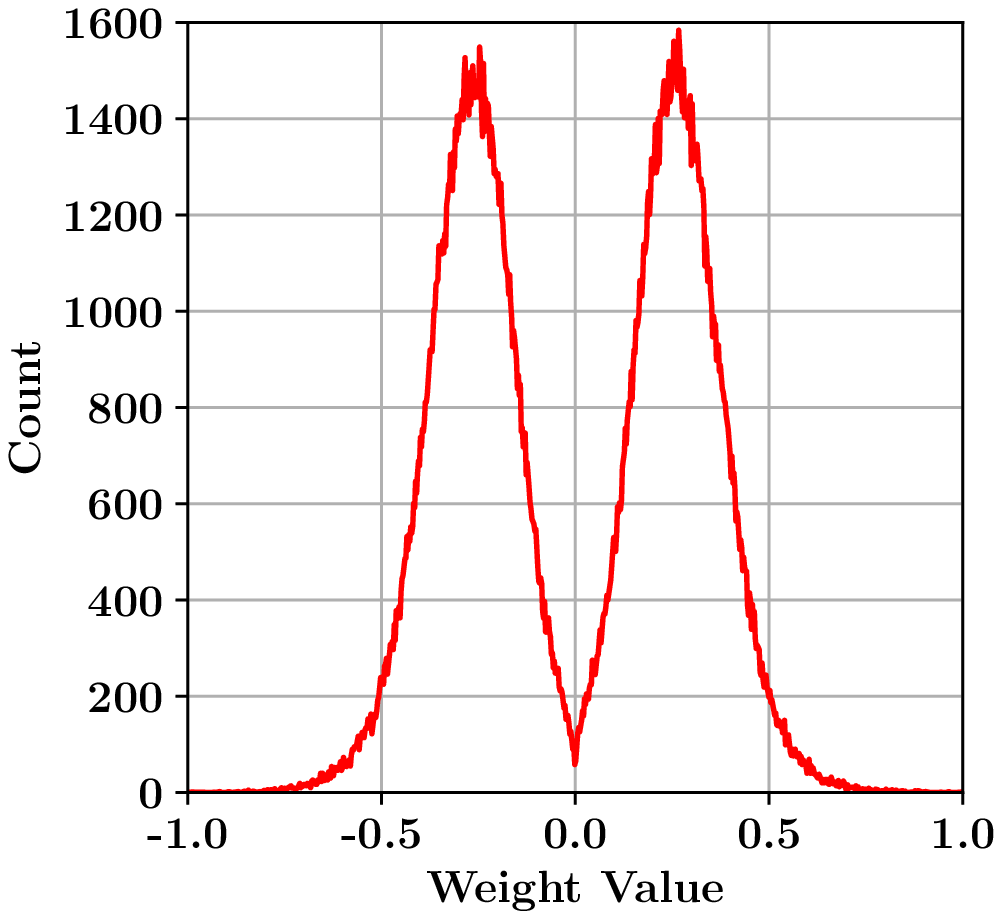}
	\end{minipage}
	\caption{Weight distribution of LSTM layer 1 of the medium PTB model after retraining with (Left) a magnitude-based pruning and (Right) $pN\!R$-based pruning with 90\% pruning rate. Our compression scheme incurs a sharp drop in the count of near-zero weights.}
	\label{fig:weight_dist_ptb}
\end{figure}

For all of the pruning rates selected, Table \ref{table:ptb_pruning} shows that our compression scheme improves perplexity better than the technique in \cite{suyog_prune} which is based on \cite{SHan_2015}.
The superiority of $pN\!R$-based pruning is partly supported by the observation that non-zero weights successfully avoid to be small through retraining while the conventional pruning still keeps near-zero (unmasked) weights as depicted in Figure~\ref{fig:weight_dist_ptb}.

\subsection{Low-Rank Approximation}

We apply our proposed occasional regularization algorithm integrated with Tucker decomposition \cite{tucker} to convolutional neural network (CNN) models and demonstrate superiority of the $pN\!R$-based scheme over conventional training methods.
In CNNs, the convolution operation requires a 4D kernel tensor $\mathcal{K}=$ $\mathbb{R}^{d \times d \times S \times T}$ where each kernel has $d\times d$ dimension, $S$ is the input feature map size, and $T$ is the output feature map size.
Then, following the Tucker decomposition algorithm, $\mathcal{K}$ is decomposed into three components as
\begin{equation}
    \label{eq:eq_tucker}
    \mathcal{\Tilde{K}}_{i,j,s,t} = \sum_{r_s=1}^{R_s}\sum_{r_t=1}^{R_t}\mathcal{C}_{i,j,r_s,r_t}\mP_{s,r_s}^S\mP_{t,r_t}^T ,
\end{equation}
where $\mathcal{C}_{i,j,r_s,r_t}$ is the reduced kernel tensor, $R_s$ is the rank for input feature map dimension, $R_t$ is the rank for output feature map dimension, and $\mP^S$ and $\mP^T$ are 2D filter matrices to map $\mathcal{C}_{i,j,r_s,r_t}$ to $\mathcal{K}_{i,j,s,t}$.
%Each component is obtained to minimize the Frobenius norm of ($\mathcal{\Tilde{K}}_{i,j,s,t}$ $-$ $\mathcal{K}_{i,j,s,t}$).
As a result, one convolution layer is divided into three convolution layers, specifically, $(1 {\times} 1)$ convolution for $\mP^S$, $(d {\times} d)$ convolution for $\mathcal{C}_{i,j,r_s,r_t}$, and $(1 {\times} 1)$ convolution for $\mP^T$ \cite{tucker_samsung}.

In prior tensor decomposition schemes, model training is performed as a fine-tuning procedure after the model is restructured and fixed \cite{cp_decomposition, tucker_samsung}.
On the other hand, our training algorithm is conducted for Tucker decomposition as follows:

\begin{enumerate}%[noitemsep,topsep=0pt,parsep=1pt,partopsep=0pt]
\item Perform normal training for $pN\!R$ (batches) without considering Tucker decomposition
\item Calculate $\mathcal{C}$, $\mP^S$, and $\mP^T$ using Tucker decomposition to obtain $\mathcal{\Tilde{K}}$
\item Replace $\mathcal{K}$ with $\mathcal{\Tilde{K}}$
\item Go to Step 1 with updated $\mathcal{K}$
\end{enumerate}
After repeating a number of the above steps towards convergence, the entire training process should stop at Step 2, and then the final decomposed structure is extracted for inference.
Because the model is not restructured except in the last step, Steps 2 and 3 can be regarded as special steps to encourage wide search space exploration so as to find a compression-friendly local minimum where weight noise by decomposition does not noticeably degrade the loss function.

Using the pre-trained ResNet-32 model with CIFAR-10 dataset \cite{resnet, tensorly}, we compare two training methods for Tucker decomposition: 1) typical training with a decomposed model and 2) $pN\!R$-based training, which maintains the original model structure and occasionally injects weight noise through decomposition.
Using an SGD optimizer, both training methods follow the same learning schedule: learning rate is 0.1 for the first 100 epochs, 0.01 for the next 50 epochs, and 0.001 for the last 50 epochs.
Except for the first layer, which is much smaller than the other layers, all convolution layers are compressed by Tucker decomposition with rank $R_s$ and $R_t$ selected to be $S$ and $T$ multiplied by a constant number $R_c$ ($0.3{\le}R_c{\le}0.7$ in this experiment).
Then, the compression ratio of a convolution layer is $d^2ST/(SR_s+d^2R_sR_t+TR_t)$ $=d^2ST/(S^2R_c+d^2R_c^2ST+T^2R_c)$, which can be approximated to be $1/R^2_c$ if $S=T$ and $d\gg R_c$.
$pN\!R$ is chosen to be 200.

\begin{figure}[t]
    \centering
    \includegraphics[width=0.48\textwidth]{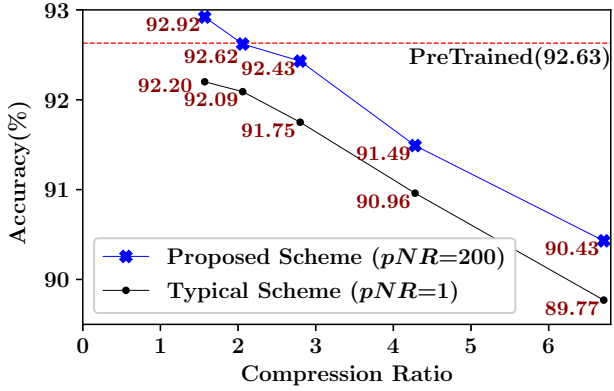}
    \caption{Test accuracy comparison on ResNet-32 using CIFAR-10 trained by typical training method and the proposed training method with various compression ratios. For the proposed scheme, test accuracy is measured only at Step 3 that allows to extract a decomposed structure, and $pN\!R$ is 200.}
    \label{fig:tucker1}
\end{figure}

\begin{figure}[t]
    \begin{minipage}{0.49\textwidth}
        \centering
	    \includegraphics[width=0.85\textwidth]{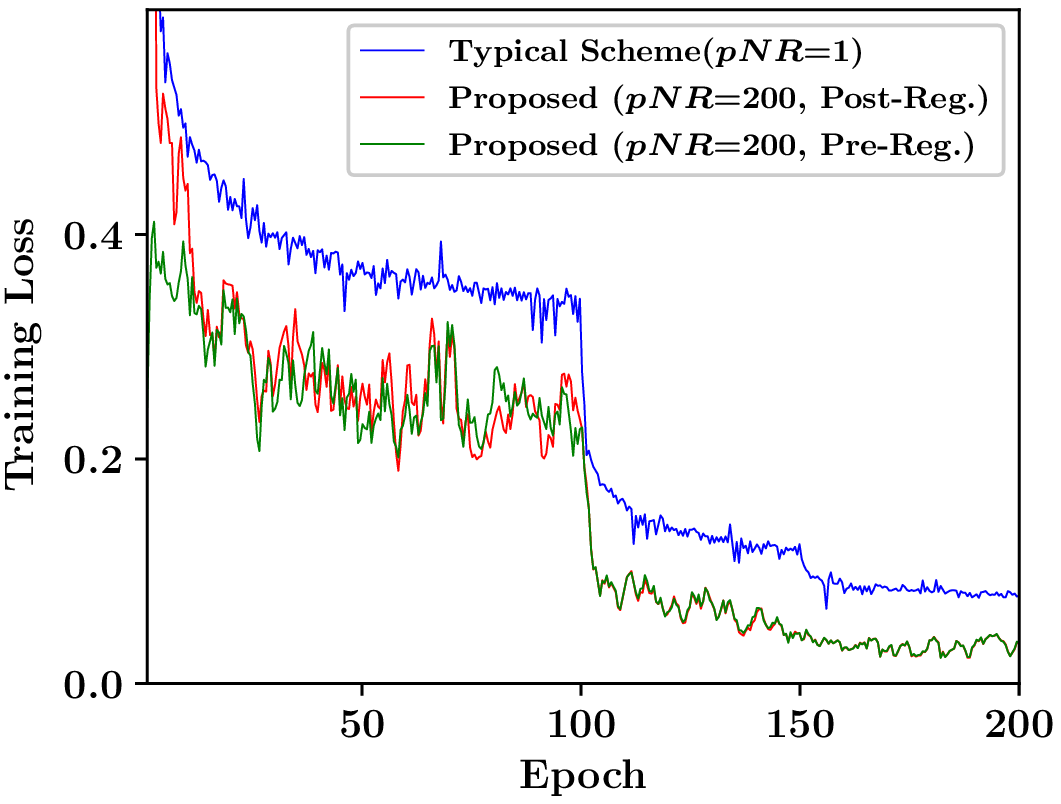}
	\end{minipage}
	\begin{minipage}{0.49\textwidth}
        \centering
	    \includegraphics[width=0.85\textwidth]{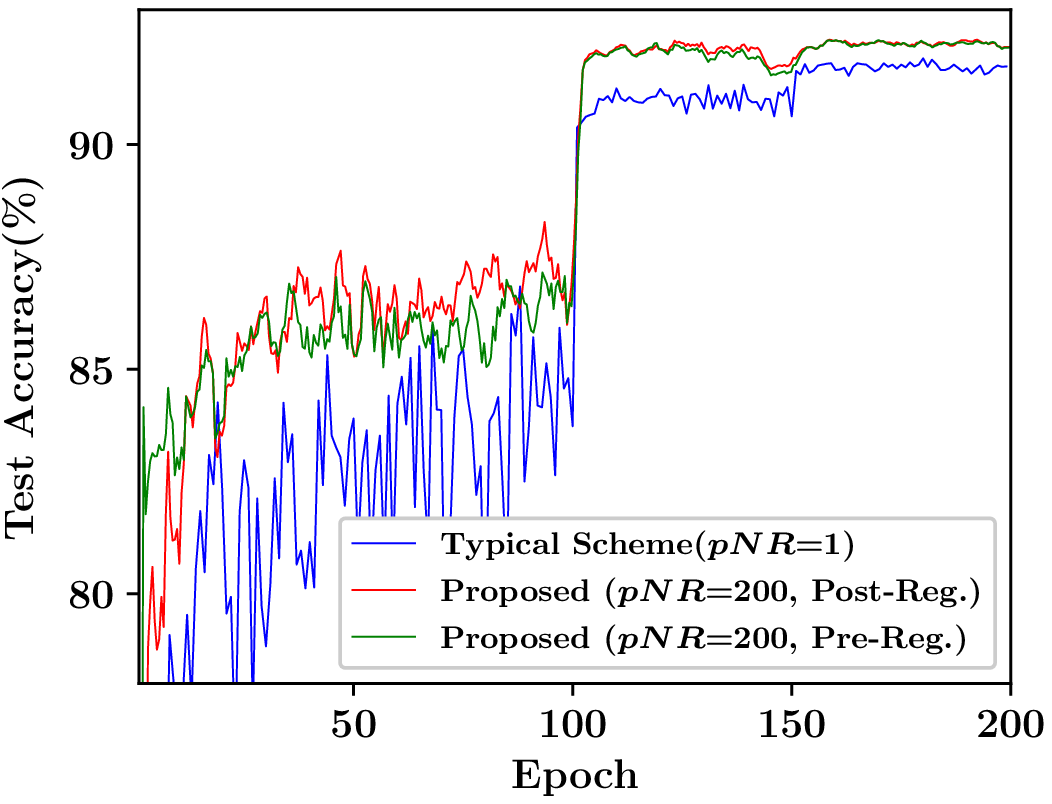}
	\end{minipage}
    \caption{Training loss (Left) and test accuracy (Right) of ResNet-32 using CIFAR-10. For the proposed scheme, training loss and test accuracy are only monitored right before or after weight regularization for compression (Pre-Reg. or Post-Reg.). Compression ratio is 2.8 with $R_c$=0.5.}
    \label{fig:tucker2}
\end{figure}

Figure~\ref{fig:tucker1} shows test accuracy after Tucker decomposition\footnote{https://github.com/larry0123du/Decompose-CNN} by two different training methods.
Note that test accuracy results are evaluated only at Step 3 where the training process can stop to generate a decomposed structure.
In Figure~\ref{fig:tucker1}, across a wide range of compression ratios (determined by $R_c$), the proposed scheme yields higher model accuracy compared to typical training.
Note that even higher model accuracy than that of the pre-trained model can be achieved by our method if the compression ratio is small enough.
In fact, Figure~\ref{fig:tucker2} shows that our technique improves training loss and test accuracy throughout the entire training process.
Initially, the gap of training loss and test accuracy between pre-regularization and post-regularization is large.
Such a gap, however, is quickly reduced through training epochs.
Overall, ResNet-32 converges successfully through the entire training process with lower training loss and higher test accuracy compared with a typical training method.

\begin{figure*}[h]
\begin{center}
\includegraphics[width=0.8\linewidth]{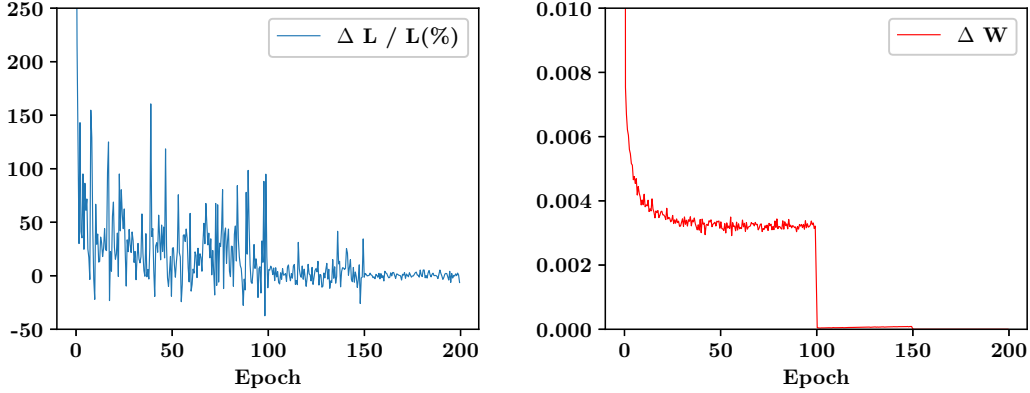}
\caption{Difference of training loss function and average Frobenius norm of weight values by weight updates for model compression. $R_c$ $=0.5$ and $pN\!R$ $=200$ are used.}
\label{fig:delta_values}
\end{center}
\end{figure*}

To investigate the effect of NR period on local minima exploration with ResNet-32 on CIFAR-10, Figure~\ref{fig:delta_values} presents the changes of loss function and weight magnitude values incurred by occasional regularization.
In Figure~\ref{fig:delta_values}(left), $\Delta \mathcal{L}/\mathcal{L}$ is given as the loss function increase $\Delta\mathcal{L}$ (due to weight regularization at $pN\!R$ steps) divided by $\mathcal{L}$, which is the loss function value right before weight regularization.
In Figure~\ref{fig:delta_values}(right), $\Delta \vw$ is defined as $||\vw - \Tilde{\vw}||^2_{\mathcal{F}}$ $/N(\vw)$, where $\vw$ is the entire set of weights to be compressed, $\Tilde{\vw}$ is the set of weights regularized by Tucker decomposition, $N(\vw)$ is the number of elements of $\vw$, and $||\mX||^2_{\mathcal{F}}$ is the Frobenius norm of $\mX$.
Initially, $\vw$ fluctuates with large corresponding $\Delta \mathcal{L}$.
Then, both $\Delta \mathcal{L}$ and $\Delta \vw$ decrease and  Figure~\ref{fig:delta_values} shows that occasional regularization finds flatter local minima (in the view of Tucker decomposition) successfully.
When the learning rate is reduced at 100th and 150th epochs, $\Delta \mathcal{L}$ and $\Delta \vw$ decrease significantly because of a lot reduced local minima exploration space.
In other words, occasional regularization helps an optimizer to detect a local minimum where Tucker decomposition does not alter the loss function value noticeably.

For ResNet-18 on ImageNet experiments and VGG19 on CIFAR-10 (including additional compression techniques), refer to Appendix.

\section{NR Period for Convergence}

\begin{figure}[h]
\centering
	\includegraphics[width=1.0\linewidth]{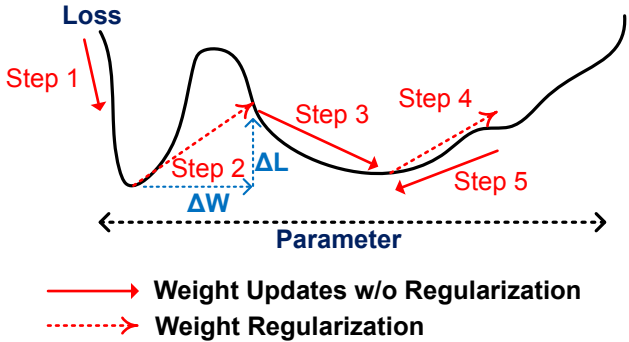}
	\caption{Gradient descent and weight regularization when NR period is given as a multiple of batches. Depending on the loss surface and/or strength of regularization, regularization would lead to step 2 (escaping from a local minimum) or step 5 (returning to a local minimum).}
	\label{fig:app_loss_surface}
\end{figure}

NR period influences convergence in training.
Strong weight regularization facilitates the chance of escaping a local minimum (depicted as step 2 in Figure~\ref{fig:app_loss_surface}) or requires longer NR period to return to a local minimum (described as step 5 in Figure~\ref{fig:app_loss_surface}).
Let us estimate the desirable NR period ($pN\!R$) considering the convergence of training even though $pN\!R$ is supposed to be searched empirically.
Given a parameter set $\vw$ (that is assumed to be close enough to a local minimum) and a learning rate $\gamma$, the loss function of a model $\mathcal{L}(\vw)$ can be approximated as 
\begin{equation}
    \label{eq:eq_loss_approxi}
    \mathcal{L}(\vw) \simeq \mathcal{L}(\vw_0) + (\vw - \vw_0)^\top (H(\vw_0)/2) (\vw - \vw_0) 
\end{equation}
using a local quadratic approximation where $H$ is the Hessian of $\mathcal{L}$ and $\vw_0$ is a set of parameters at a local minimum.
After regularization is performed at step $t$, $\vw$ can be updated by gradient descent as follows:
\begin{equation}
    \label{eq:eq_1}
    \vw_{t+1} = \vw_{t} - \gamma \frac{\partial \mathcal{L}}{\partial \vw}\big|_{\vw=\vw_t}  \simeq \vw_{t} - \gamma H(\vw_0)(\vw_t - \vw_0).
\end{equation}
Thus, after $pN\!R$, we obtain
\begin{equation}
    \label{eq:eq_w_pNR}
    \vw_{t+pN\!R} = \vw_0 + (I-\gamma H(\vw_0))^{pN\!R} (\vw_t - \vw_0),
\end{equation}
where $I$ is an identity matrix.
Suppose that $H$ is positive semi-definite and all elements of $I - \gamma H(\vw_0)$ are less than 1.0, $\vw_{t+pN\!R}$ can converge to $\vw_0$ with long $pN\!R$ which should be longer with larger $(\vw_t - \vw_0)$ (i.e., stronger weight regularization) or smaller $\gamma H(\vw_0)$.

\section{Related Work}

Periodic compression has been introduced in the literature to gradually improve compression ratio or automate hyper-parameter search process.
DropPruning repeats dropping weights randomly and retraining the model while some previously dropped weights are unpruned until pruning rate reaches a target number \cite{jia2018droppruning}.
Weights are incrementally quantized to improve model accuracy \cite{zhou2017incremental} or the number of quantization bits can be controlled differently for each layer by a loop based on reinforcement learning \cite{elthakeb2018releq}.
Structured pruning and fine-tuning process can be iterated to increase pruning rate \cite{molchanov2016pruning, liu2017learning}.
All of these previous works assume $pN\!R=1$ (i.e., performing compression for every mini batch) while the goal is increasing compression ratio slowly or finding a set of hyper-parameters through iterative fine-tuning stages.
Our proposed compression technique can be combined with such periodic compression methods (incremental compression or automatic hyper-parameter selection are also applicable to our proposed method).
In the work by \cite{he2018soft}, soft filter pruning is conducted with $pN\!R$ = 1 epoch without analysis of why such occasional pruning improves model accuracy.

\section{Conclusion}

In this paper, we introduce a new hyper-parameter called non-regularization period or NR period during which weights are updated only for gradient computations.
NR period (or equivalently regularization frequency) provides a critical impact on the overall regularization strength.
For example, if a weight decay factor becomes larger, then NR period can be longer to maintain the regularization strength.
Using such a property, we demonstrate that during compression-aware training, NR period can control the regularization strength given a target compression ratio such that model accuracy is improved compared to the case of compression for every mini-batch.
Throughout various experiments, we show that there is a particular NR period (associated with occasional weight compression accordingly) that maximizes model accuracy. 
%Similar to learning rate or dropout rate, NR period is a hyper-parameter that needs to be empirically explored. 

% % Acknowledgements should only appear in the accepted version.
% \section*{Acknowledgements}

% \textbf{Do not} include acknowledgements in the initial version of
% the paper submitted for blind review.

% If a paper is accepted, the final camera-ready version can (and
% probably should) include acknowledgements. In this case, please
% place such acknowledgements in an unnumbered section at the
% end of the paper. Typically, this will include thanks to reviewers
% who gave useful comments, to colleagues who contributed to the ideas,
% and to funding agencies and corporate sponsors that provided financial

\nocite{langley00}

\bibliography{MLsys20_DeepTwist}
\bibliographystyle{mlsys2020}

\appendix
\clearpage

\section{Weight Decay and Weight Noise Insertion}

Weight decay is one of the most well-known regularization techniques \cite{three_mecha} and different from $L_2$ regularization in a sense that weight decay is separated from the loss function calculation \cite{decoupledweightdecay}.
Weight decay is performed as
\begin{equation}
\label{eq:weight_decay}
    \vw_{t+1} = (1 - \gamma \theta \vw_t) - \gamma \nabla_{\vw_t} \mathcal{L}(\vw),
\end{equation}
where $\theta$ is a constant weight decay factor.
Weight noise insertion is another regularization technique aiminig at reaching flat minima \cite{deeplearningbook, flat_minima}.
Suppose that random Gaussian noise is added to weights such that $\vw' \!= \!\vw \!+ \!\bm{\epsilon}$ when $\bm{\epsilon} \!\sim\! \mathcal{N}(0,\eta I)$.
Then, $\mathcal{L}(\vw')$ = $\E [f_{\vw+\bm{\epsilon}} (\vx) -\vy]^2$ where $\vx$, $\vy$, $f$ are input, target, and prediction function, respectively.
Using Taylor-series expansion to second-order terms, we obtain $f_{\vw+\bm{\epsilon}}(\vx) \approx f_{\vw}(\vx) + \bm{\epsilon}^\top \nabla f(\vx) + \bm{\epsilon}^\top \nabla^2 f(\vx)\bm{\epsilon} /2$.
Correspondingly, the loss function can also be approximated as
\begin{multline}
    \label{eq:loss_weight_noise}
    \mathcal{L}(\vw+\bm{\epsilon}) \approx \E[f_\vw (\vx) - \vy]^2 \\ + \eta \E[(f_\vw (\vx) - \vy) \nabla^2 f_\vw (\vx)] + \eta \E || \nabla f_\vw (\vx) ||^2,
\end{multline}
where the second term disappears near a local minimum and the third term induces flat minima.
Random noise insertion with other distribution models can be explained in a similar fashion \cite{deeplearningbook}.

\section{Supplementary Experiments for Weight Decay and Weight Noise}

\begin{figure}[t]
    \centering
	\begin{minipage}[t]{0.49\textwidth}
		\includegraphics[width=1\linewidth]{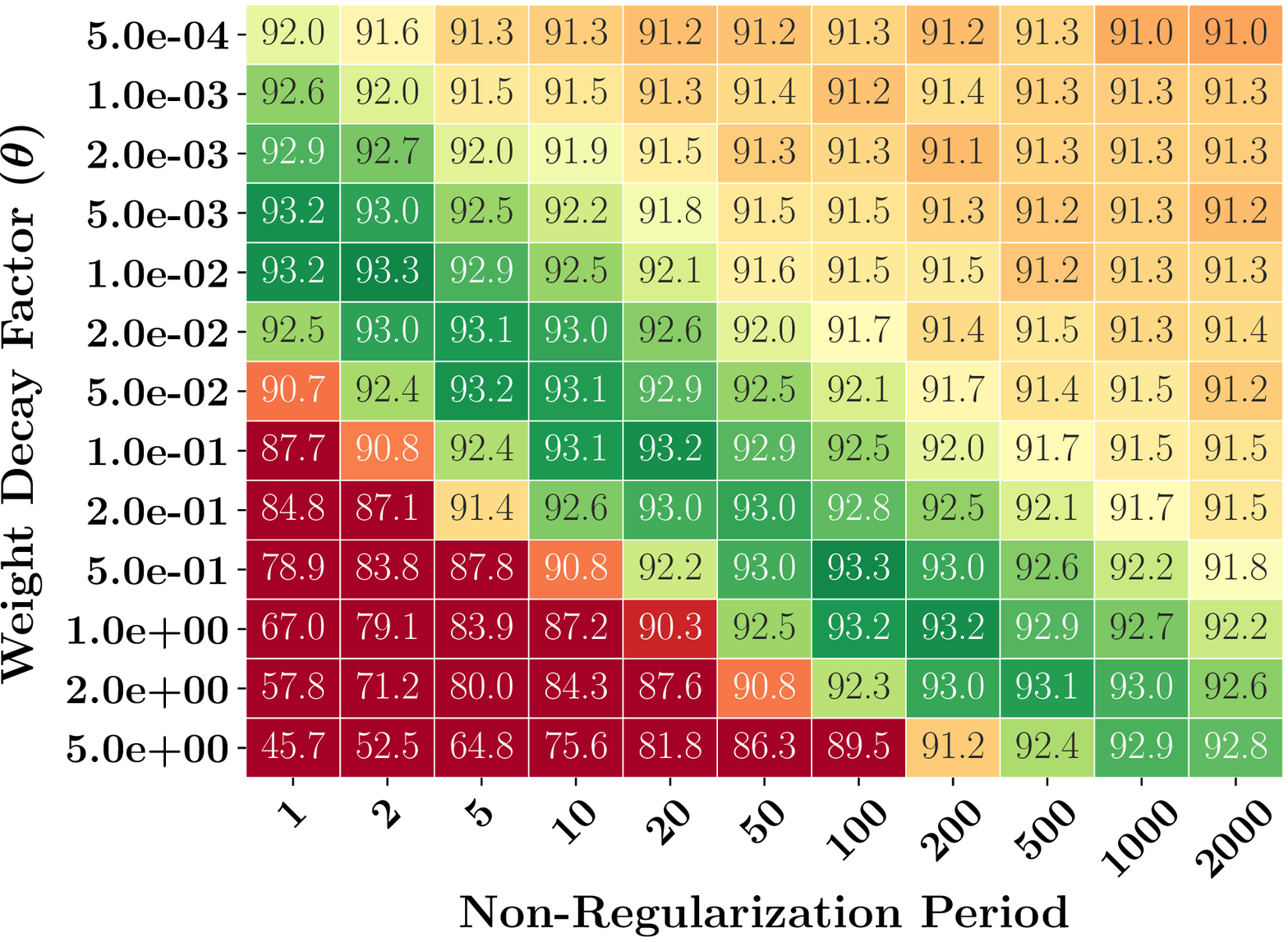}
	\end{minipage}
	\begin{minipage}[t]{0.49\textwidth}
		\includegraphics[width=1\linewidth]{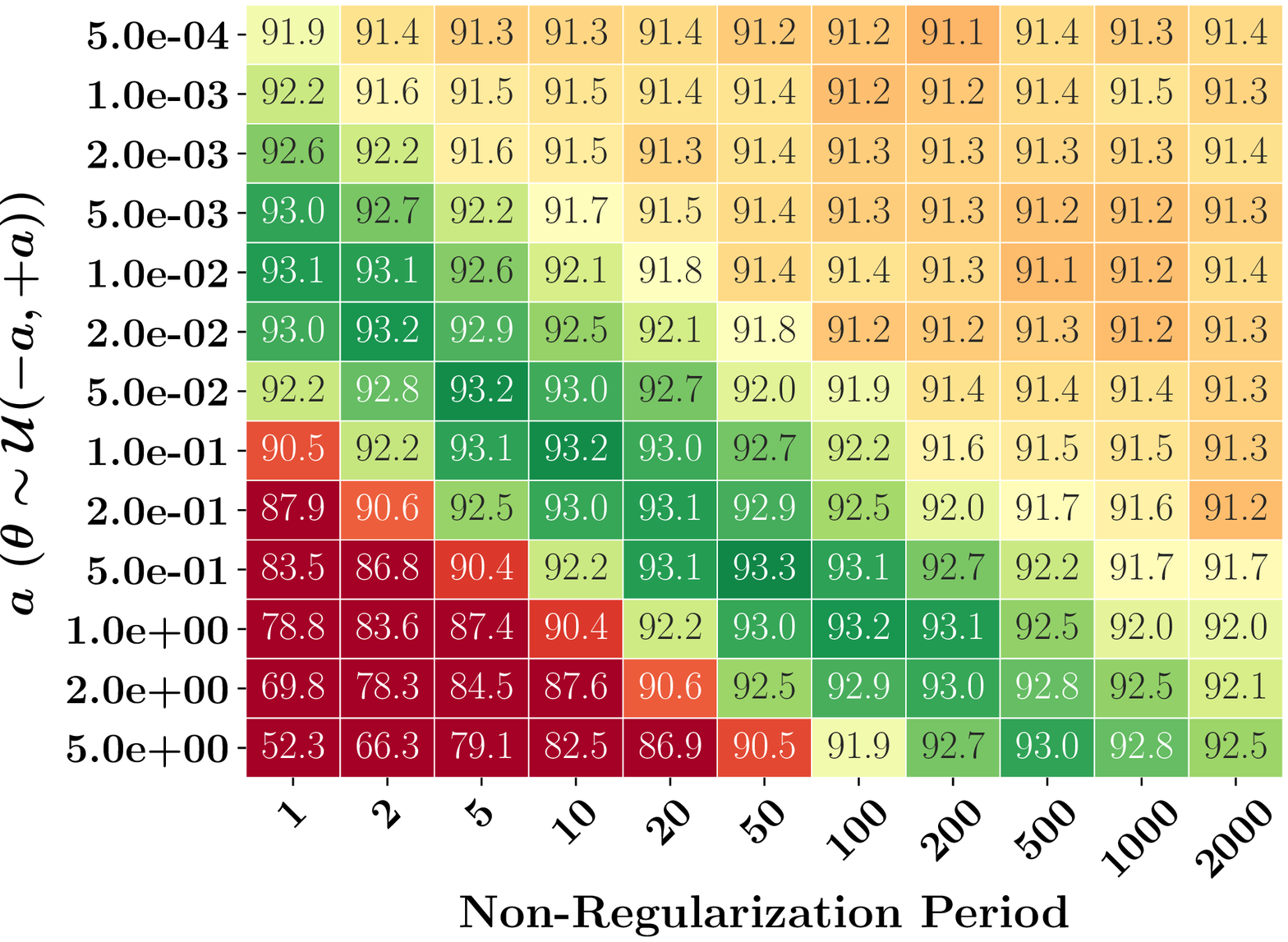}
	\end{minipage}
	\caption{Model accuracy of ResNet-32 on CIFAR-10 using various NR period and amount of weight regularization (original model accuracy without regularization is 92.6\%). (Top): Weight decay. (Bottom): Uniform weight noise insertion.}
	\label{fig:app_resnet32_noise}
\end{figure}

\begin{table*}[h]
\caption{Model accuracy of ResNet-32 on CIFAR-10 and LSTM model on PTB with various weight decay factor and corresponding $pN\!R$.}
\label{table:optimal_pNR_weight_decay}
\begin{center}
\begin{tabular}{c c | c c c c c c c}
\Xhline{2\arrayrulewidth}
& & \multicolumn{7}{c}{Weight Decay Factor($\theta)$} \\
Model & & 0 & 1e-4 & 5e-4 & 1e-3 & 5e-3 & 1e-2 & 5e-2 \\
\Xhline{2\arrayrulewidth}
\mr{2}{ResNet-32} & Accuracy(\%) & 92.6 & 93.3 & 93.2 & 93.2 & 93.3 & 93.2 & 92.9 \\
\cline{2-9}
& {optimal $pN\!R$} & N/A & 2 & 5 & 20 & 100 & 200 & 1000\\ 
\hline
\mr{2}{LSTM on PTB} & Perplexity & 114.6 & 108.1 & 97.7 & 97.1 & 97.1 & 97.0 & 97.2 \\
\cline{2-9}
& {optimal $pN\!R$} & N/A & 1 & 1 & 1 & 5 & 10 & 100 \\
%ResNet32 & Weight Decay & 92.6 & 90.0 (pNR=4) & & \\
%ResNet32 & Uniform Noise & 92.6 & 90.0 (pNR=4) & & \\
%PTB & Weight Decay & 114.6 & 97.1(pNR=1) & 97.0 (pNR=10) & 97.0(pNR=154)\\
%PTB & Uniform Noise & 114.6 & 114.4(pNR=1) & 113.0 (pNR=2) & 113.6(pNR=464) \\
\Xhline{2\arrayrulewidth}
\end{tabular}
\end{center}
\end{table*}

\begin{figure}[h]
\centering
	\includegraphics[width=1.0\linewidth]{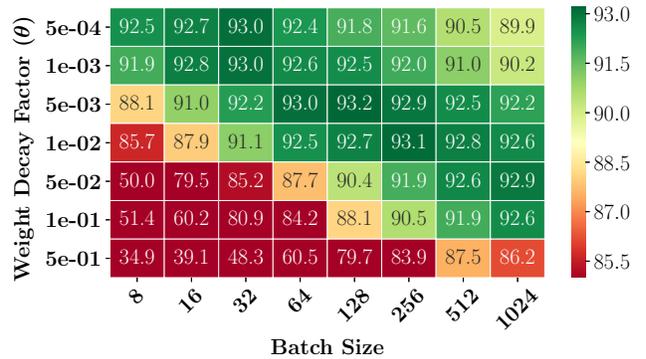}
	\caption{Model accuracy (\%) of ResNet-32 for various weight decay factors and batch size when $pN\!R$=$1$. Large batch size demands larger weight decay factors that is also reported by \cite{decoupledweightdecay}.}
	\label{fig:res_batch}
\end{figure}

\begin{figure*}[h]
    \begin{center}
	\begin{subfigure}[t]{.75\textwidth}
	    \centering
		\includegraphics[width=1\linewidth]{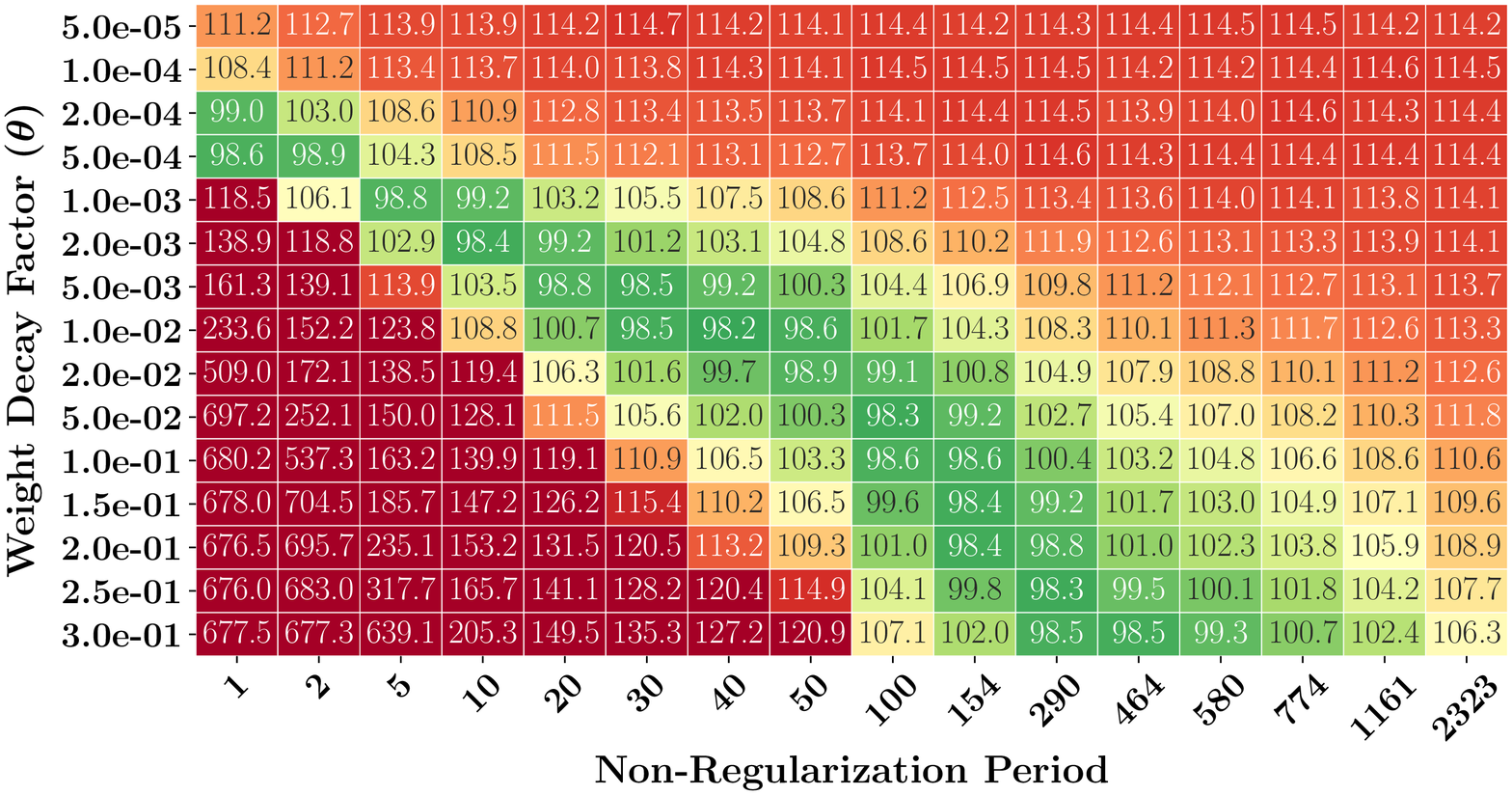}
		\caption{Initial learning rate = 1.0}
	\end{subfigure}
	
	\begin{subfigure}[t]{.75\textwidth}
	\centering
		\includegraphics[width=1\linewidth]{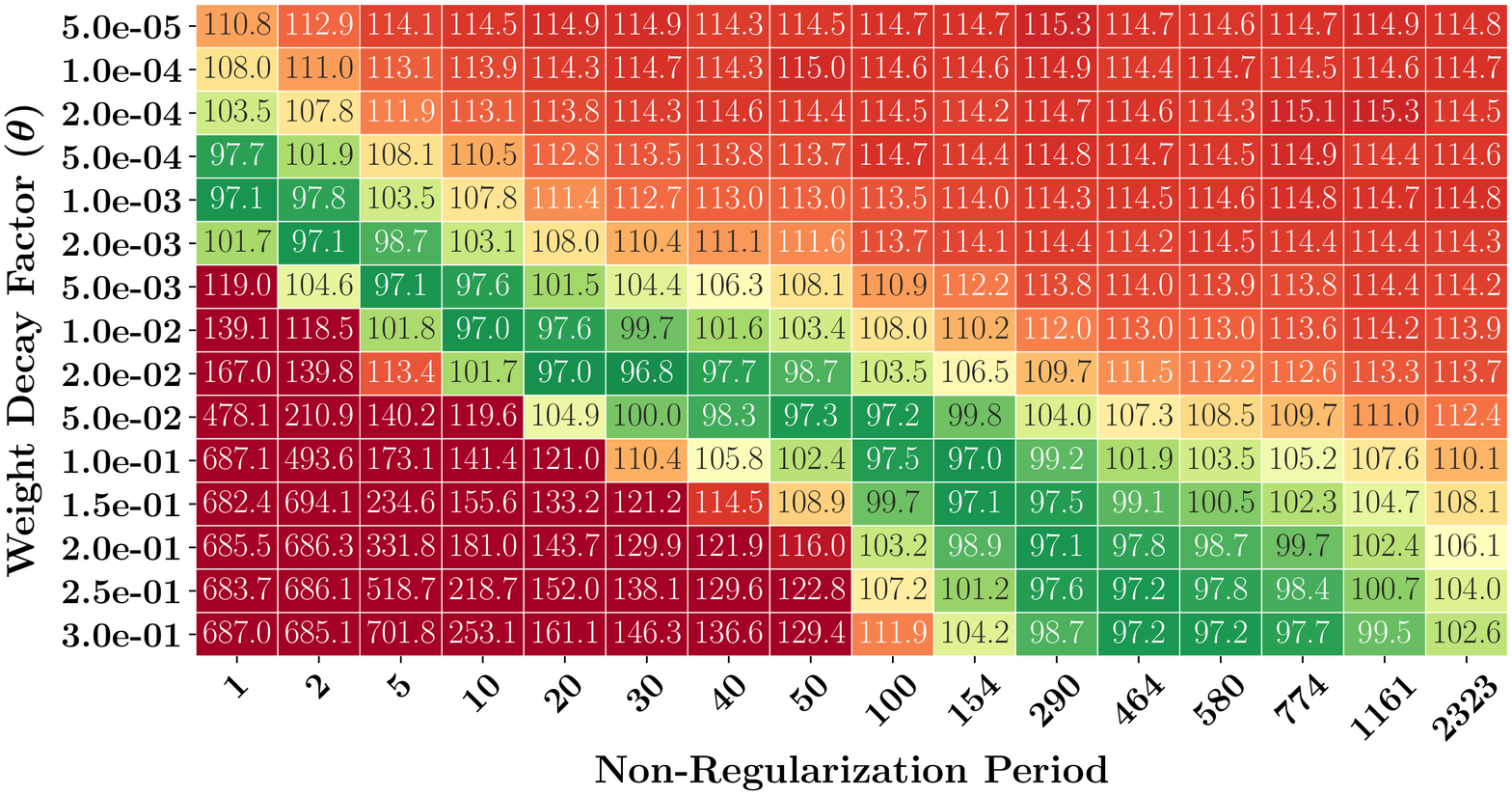}
		\caption{Initial learning rate = 1.5}
	\end{subfigure}
	
	\begin{subfigure}[t]{.75\textwidth}
	\centering
		\includegraphics[width=1\linewidth]{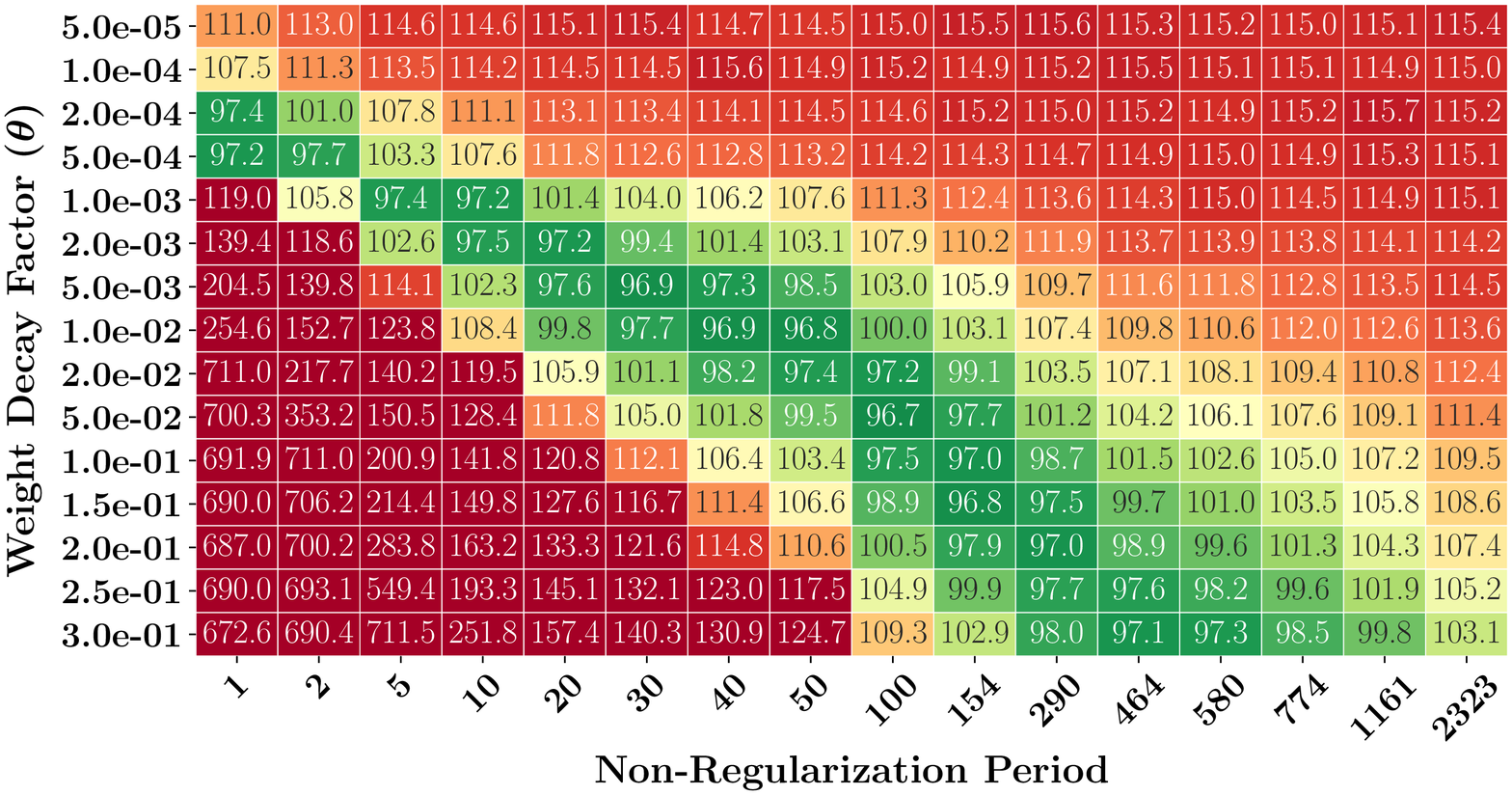}
		\caption{Initial learning rate = 2.0}
	\end{subfigure}
	
	\caption{Perplexity of LSTM model on PTB dataset using various NR period and amounts of weight decay (original perplexity without regularization is 114.60).}
	\label{fig:app_ptb_noise_wd}
	\end{center}
\end{figure*}

\begin{figure*}[h]
    \begin{center}
	\begin{subfigure}[t]{.75\textwidth}
	    \centering
		\includegraphics[width=1\linewidth]{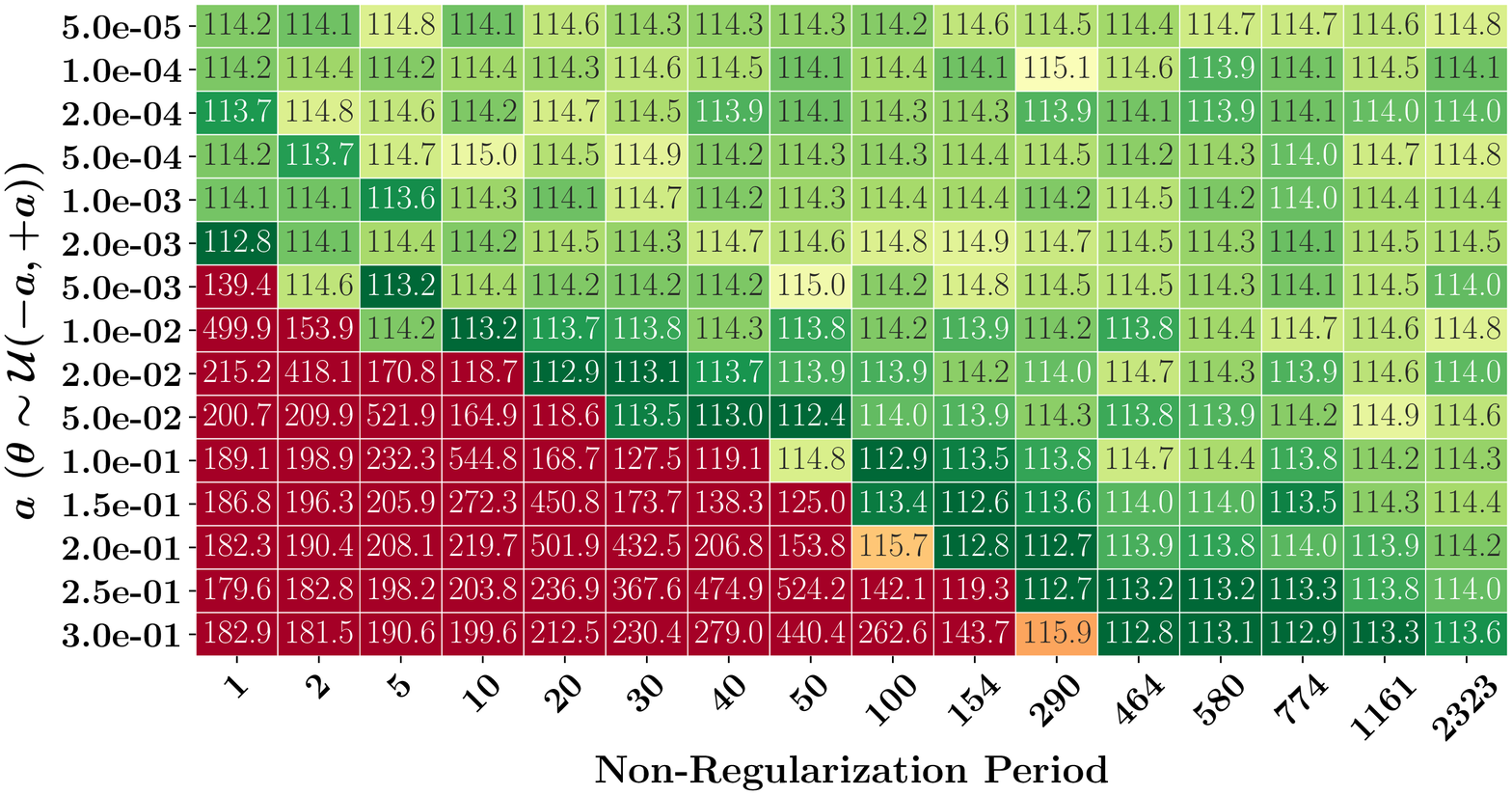}
		\caption{Initial learning rate = 1.0}
	\end{subfigure}
	
	\begin{subfigure}[t]{.75\textwidth}
	\centering
		\includegraphics[width=1\linewidth]{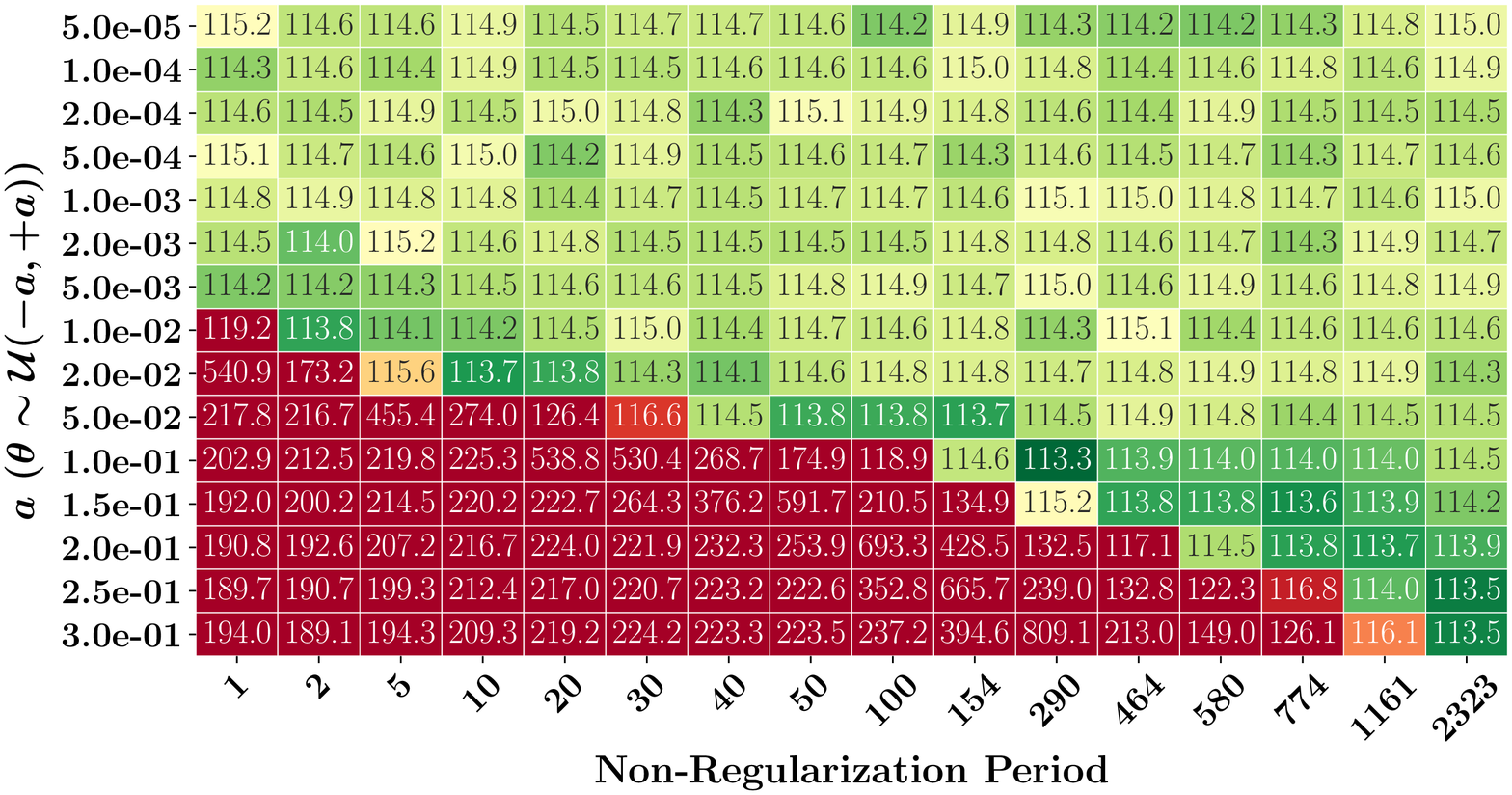}
		\caption{Initial learning rate = 1.5}
	\end{subfigure}
	
	\begin{subfigure}[t]{.75\textwidth}
	\centering
		\includegraphics[width=1\linewidth]{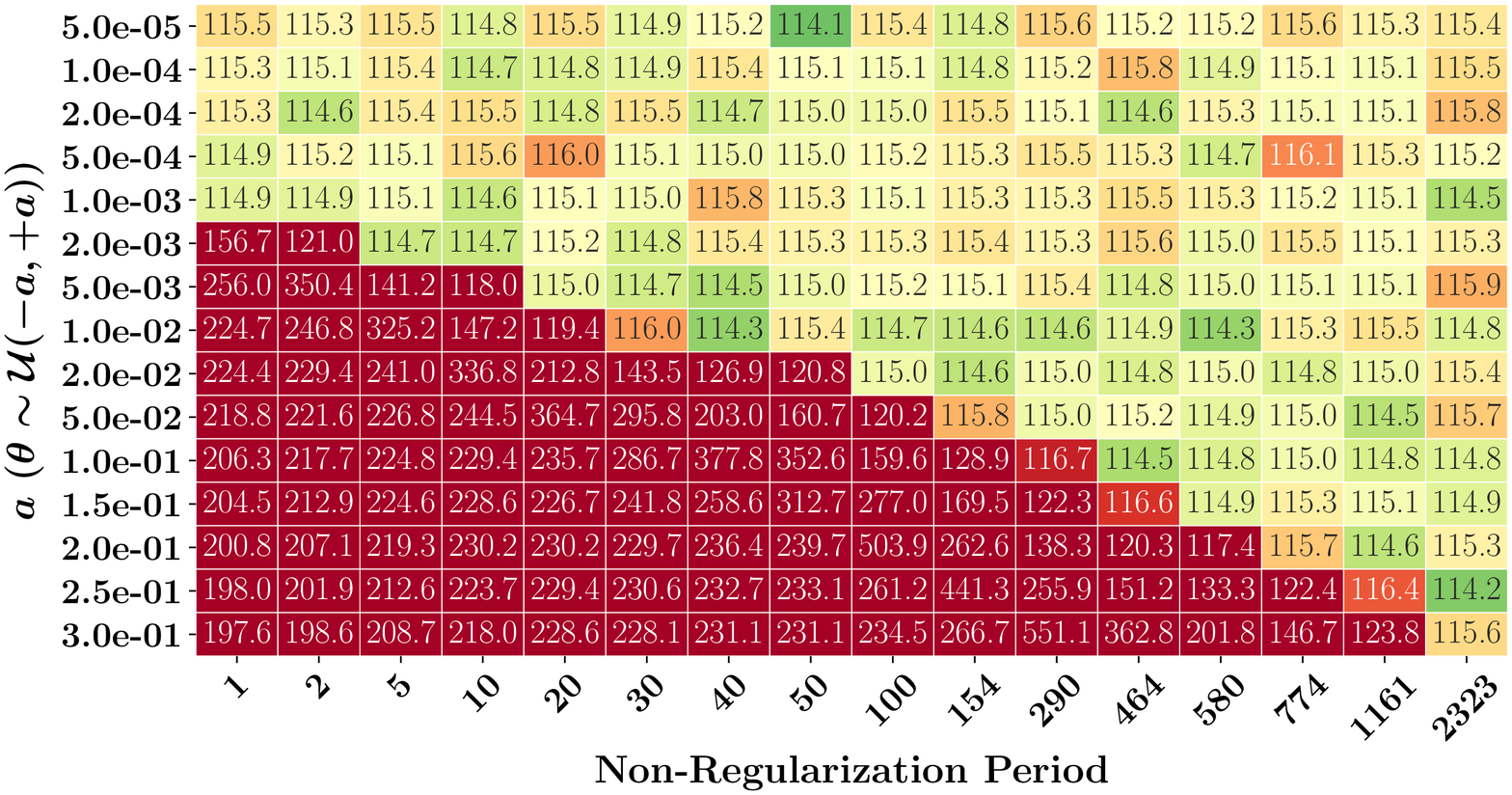}
		\caption{Initial learning rate = 2.0}
	\end{subfigure}
	
	\caption{Perplexity of LSTM model on PTB dataset using various NR period and amounts of uniform noise (original perplexity without regularization is 114.60).}
	\label{fig:app_ptb_noise_uniform}
    \end{center}
\end{figure*}

\begin{figure*}
    \begin{center}
	\begin{subfigure}[t]{1.0\textwidth}
		\includegraphics[width=1\linewidth]{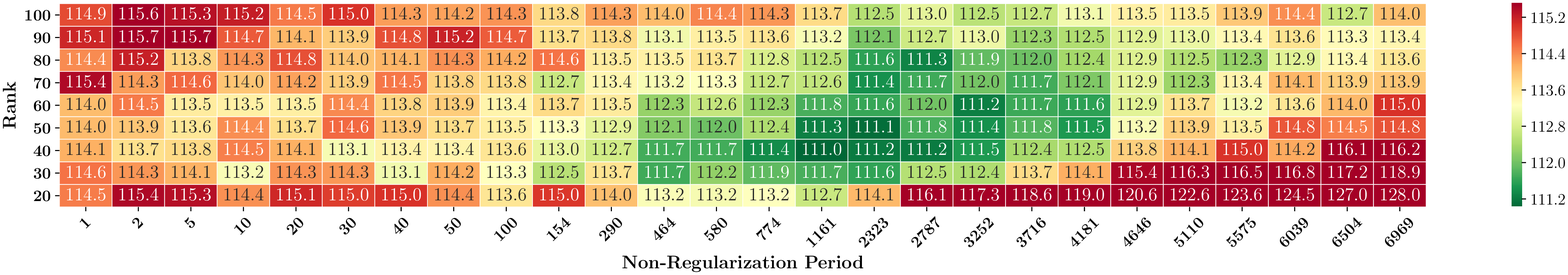}
		\caption{SVD}
	\end{subfigure}
	
    \begin{subfigure}[t]{1.0\textwidth}
		\includegraphics[width=1\linewidth]{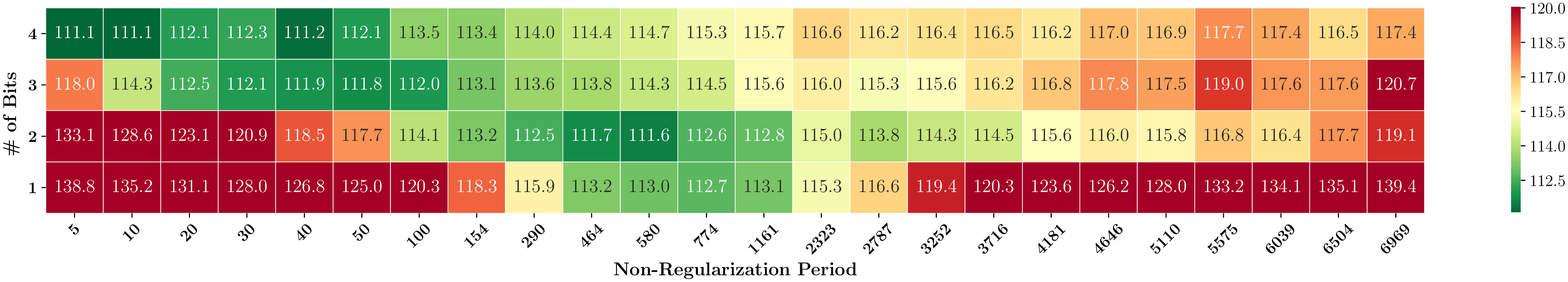}
		\caption{Quantization}
	\end{subfigure}
	
	\begin{subfigure}[t]{1.0\textwidth}
		\includegraphics[width=1\linewidth]{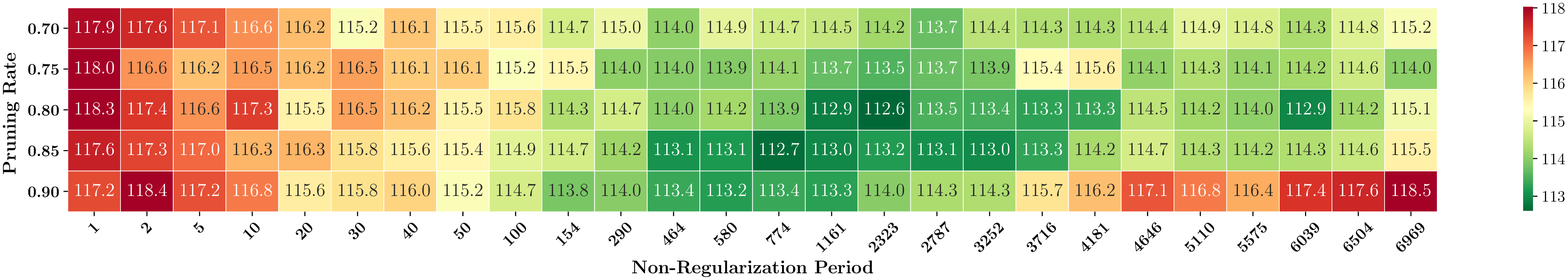}
			\caption{Pruning}
	\end{subfigure}
	
	\caption{Relationship between test perplexity and $pN\!R$ using PTB LSTM model.}
	\label{fig:app_ptb_comp}
    \end{center}
\end{figure*}

\clearpage

\section{2-Dimensional SVD Enabled by Occasional Regularization}

In this subsection, we discuss why 2D SVD needs to be investigated for CNNs and how occasional regularization enables a training process for 2D SVD.

\subsection{Issues of 2D SVD on Convolution Layers}

Convolution can be performed by matrix multiplication if an input matrix is transformed into a Toeplitz matrix with redundancy and a weight kernel is reshaped into a $T \times (S \times d \times d)$ matrix (i.e., a lowered matrix) \cite{lowering}.
Then, commodity computing systems (such as CPUs and GPUs) can use libraries such as Basic Linear Algebra Subroutines (BLAS) without dedicated hardware resources for convolution \cite{minsik_mec}.
Some recently developed DNN accelerators, such as Google's Tensor Processing Unit (TPU) \cite{TPU}, are also focused on matrix multiplication acceleration (usually with reduced precision).

For BLAS-based CNN inference, reshaping a 4D tensor $\mathcal{K}$ and performing SVD is preferred for low-rank approximation rather than relatively inefficient Tucker decomposition followed by a lowering technique. 
However, a critical problem with SVD (with a lowered matrix) for convolution layers is that two decomposed matrices by SVD do not present corresponding (decomposed) convolution layers, because of intermediate lowering steps.
As a result, fine-tuning methods requiring a structurally modified model for training are not available for convolution layers to be compressed by SVD.
On the other hand, occasional regularization does not alter the model structure for training.
For occasional regularization, SVD can be performed as a way to feed noise into a weight kernel $\mathcal{K}$ for every regularization step.
Once training stops at a regularization step, the final weight values can be decomposed by SVD and used for inference with reduced memory footprint and computations.
In other words, occasional regularization enables SVD-aware training for CNNs.

\begin{figure*}[h]
\hfill
	\begin{minipage}[t]{.54\textwidth}
		\includegraphics[width=1\linewidth]{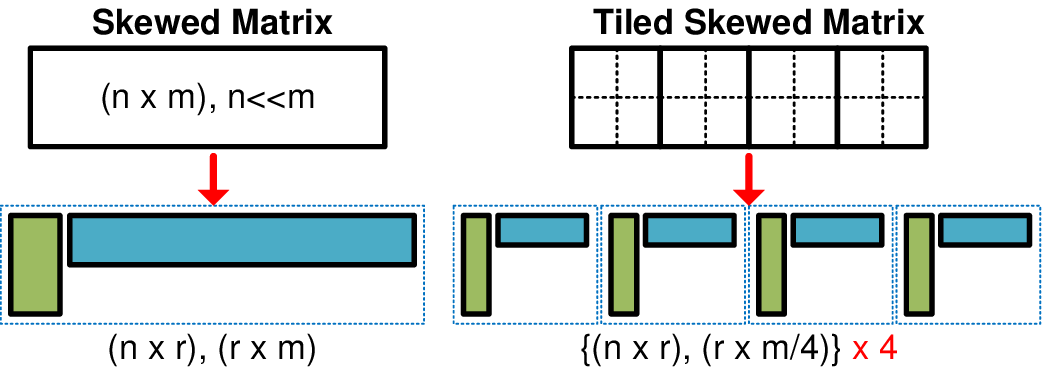}
	\end{minipage}
	\hfill
	\begin{minipage}[t]{.44\textwidth}
		\includegraphics[width=1\linewidth]{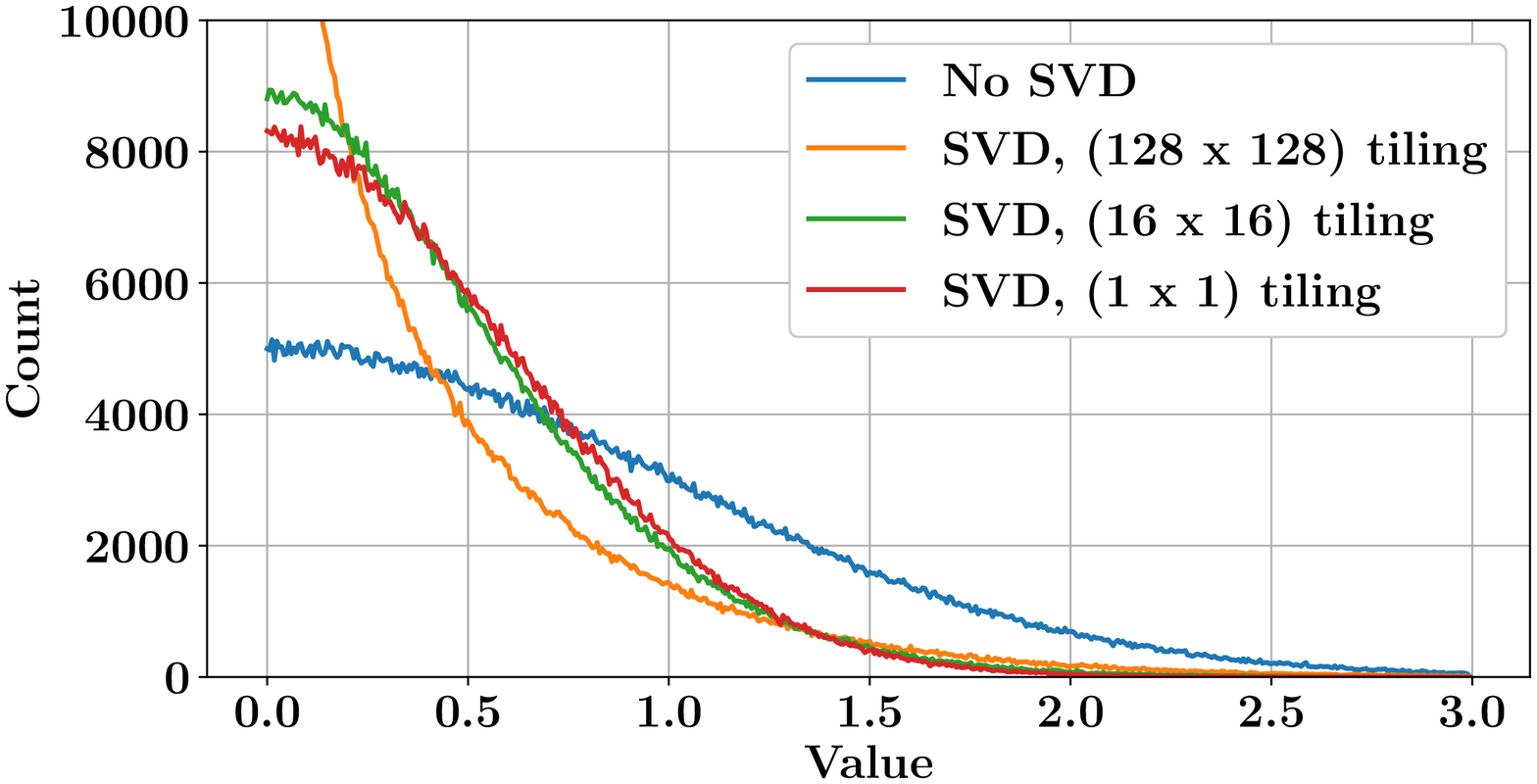}
	\end{minipage}
	\hfill
	\caption{Skewed matrix and a tiling technique are illustrated on the left side, while the right side presents distributions of weights after SVD with different tiling schemes (only positive weights are included).}
	\label{fig:nmf_tiling2}
\end{figure*}

\subsection{Tiling-Based SVD for Skewed Weight Matrices}

A reshaped kernel matrix $\mK$ $\in\R^{T\times (S \times d \times d)}$ is usually a skewed matrix where row-wise dimension ($n$) is smaller than column-wise dimension ($m$) as shown in Figure~\ref{fig:nmf_tiling2} (i.e., $n \ll m$).
A range of available rank $r$ for SVD, then, is constrained by small $n$ and the compression ratio is approximated to be $n/r$.
If such a skewed matrix is divided into four tiles as shown in Figure~\ref{fig:nmf_tiling2} and the four tiles do not share much common chateracteristics, then tiling-based SVD can be a better approximator and rank $r$ can be further reduced without increasing approximation error.
Moreover, fast matrix multiplication is usually implemented by a tiling technique in hardware to improve the weight reuse rate \cite{tiling_matmul}.
Hence, tiling could be a natural choice not only for high-quality SVD but also for high-performance hardware operations.

To investigate the impact of tiling on weight distributions after SVD, we tested a $(1024\times1024)$ random weight matrix in which elements follow a Gaussian distribution.
A weight matrix is divided by $(1\times1)$, $(16\times16)$, or $(128\times128)$ tiles (then, each tile is a submatrix of $(1024\times1024)$, $(64\times64)$, or $(8\times8)$ size).
Each tile is compressed by SVD to achieve the same overall compression ratio of $4\times$ for all of the three cases.
As described in Figure~\ref{fig:nmf_tiling2} (on the right side), increasing the number of tiles tends to increase the count of near-zero and large weights (i.e., variance of weight values increases).
Figure~\ref{fig:nmf_tiling2} can be explained by sampling theory where decreasing the number of random samples (of small tile size) increases the variance of sample mean.
In short, tiling affects the variance of weights after SVD (while the impact of such variance on model accuracy should be empirically studied).

%ResNet32 + CIFAR10 실험에서 64x64x3x3 Layer 9개만 압축
%2,4배 압축율에 대해서 tile size 변화에 따른 200epoch test accuracy 측정
%기타 실험 조건은 동일. (dist=200)
\begin{table*}[h]
\begin{center}
\caption{Test accuracy(\%) of ResNet-32 model using CIFAR-10 dataset while the 9 largest convolution layers ($T$=$S$=64, $d$=3) are compressed by SVD using different tiling configurations. For each tile size, rank $r$ is selected to achieve compression ratio of $2\times$ or $4\times$. $pN\!R$=200 is used for occasional regularization.} 
\label{table:tileacc}
\begin{tabular}{c||c|ccccc}
\Xhline{2\arrayrulewidth}
            \mr{2}{Pre-Trained} & Compression & \multicolumn{4}{c}{\begin{tabular}[c]{@{}c@{}}Size of Each Tile\end{tabular}} \\
            \cline{3-6}
            & Ratio & 64$\times$64 & 32$\times$32 & 16$\times$16 & 8$\times$8   \\
\Xhline{2\arrayrulewidth}
 \mr{2}{92.63} & 2$\times$    & 93.34 ($r$=16) & 93.11 ($r$=8) & 93.01 ($r$=4) & 93.23 ($r$=2) \\
 & 4$\times$   & 92.94 ($r$=8)~~ & 92.97 ($r$=4) & 93.00 ($r$=2) & 92.81 ($r$=1) \\
\Xhline{2\arrayrulewidth}
\end{tabular}
\end{center}
\end{table*}

We applied the tiling technique and SVD to the 9 largest convolution layers of ResNet-32 using the CIFAR-10 dataset.
Weights of selected layers are reshaped into $64\times(64\times3\times3)$ matrices with the tiling configurations described in Table~\ref{table:tileacc}.
We perform training with the same learning schedule and $pN\!R$(=200) used in Section 3.
Compared to the test accuracy of the pre-trained model (=92.63\%), all of the compressed models in Table~\ref{table:tileacc} achieves higher model accuracy due to the regularization effect of our compression scheme.
Note that for each target compression ratio, the relationship between tile size and model accuracy is not clear.
Hence, various configurations of tile size need to be explored to enhance model accuracy, even though variation of model accuracy for different tile size is small.

% Graph : TiledSVD + SVD 한 결과 여러개
% ResNet32 + CIFAR10 실험 공통 : 
%  - ResNet32 + CIFAR10 (PreTrained : 92.63)
%  - 16x3x3x3(첫 Layer) 제외하고 모든 Layer 동일한 압축율로 압축
%  - 200epoch(lr0.1 100epoch, lr 0.01 50epoch, lr 0.001 50epoch)
% DeepTwist : Distortion Step = 200 (모두 200으로 수정)
%  - Tiled SVD with (16,16,16) = 16x16 fixed tile로 모든 layer를 조각냄.
%  - Tiled SVD with (16,32,32) = 32x32 fixed tile로 조각내기에는 16x16x3x3 conv를 조각낼수 없기 때문에 16x16x3x3, 32x16x3x3 conv layer는 16x16 tile로 조각내고, 나머지는 32x32로 조각냄.

\begin{figure*}[h]
\begin{center}
    \centering
    \includegraphics[width=0.8\linewidth]{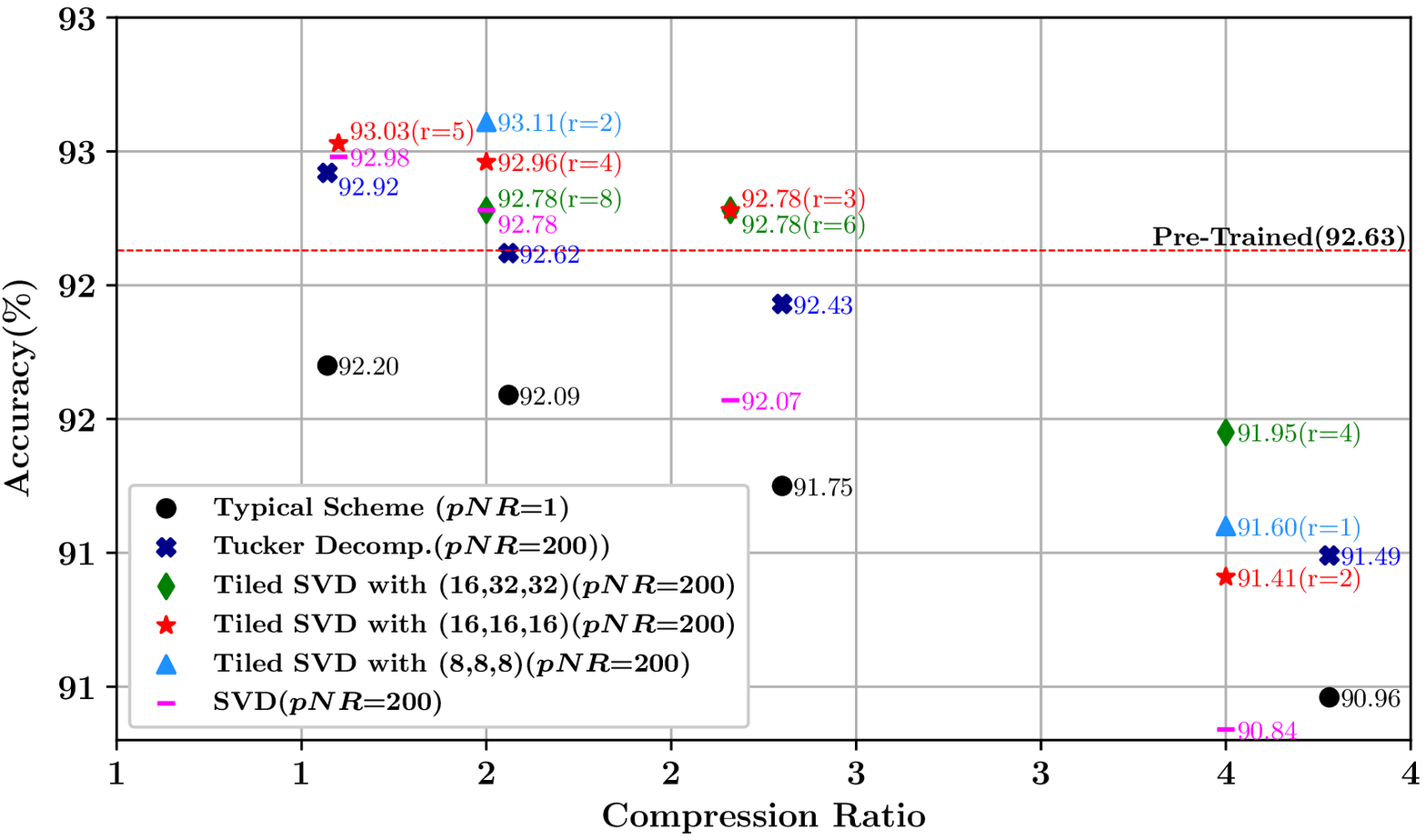}
\caption{Test accuracy of ResNet-32 model using CIFAR-10 with various target compression ratio and decomposition methods. Except the first small convolution layer, all layers are compressed by the same compression ratio. Convolution layers can be grouped according to 3 different $S$ values (=16, 32, or 64). For tiled SVD, three groups (of different $S$) are tiled in ($k_1 \times k_1$), ($k_2 \times k_2$), or ($k_3 \times k_3)$ tile size. ($k_1, k_2, k_3$) configuration is described in legends.  } 
\label{fig:svd}
\end{center}
\end{figure*}

\section{Experimental Results on Low-Rank Approximation for CNNs}

In this subsection, we apply low-rank approximation trained by occasional regularization to various CNN models.

Figure~\ref{fig:svd} summarizes the test accuracy values of ResNet-32 (with CIFAR-10 dataset) compressed by various low-rank approximation techniques.
Note that tiled SVD and normal SVD are enabled only by occasional regularization, which obviates model structure modification during training.
All configurations in Figure~\ref{fig:svd} use the same learning rate scheduling and the number of training epochs as described in Section 3.
Results show that tiled SVD yields the best test accuracy and test accuracy is not highly sensitive to tile configuration.
SVD presents competitive model accuracy for small compression ratios.
As compression ratio increases, however, model accuracy using SVD significantly degrades.
From Figure~\ref{fig:svd}, tiled SVD associated with occasional regularization is clearly the best low-rank approximation scheme.

% Table : VGG19 + CIFAR10에 대한 Tucker vs TiledSVD Test Accuracy 비교 (2배, 2/4배, 4배) // DeepTwist Tucker는 없음.
% 1. Pre-Trained Model : 92.37 (lr 0.1부터 50epoch마다 1/2 decay, 300epoch)
% 2. Fully-connected layer(512x512) 두개를 8x 압축해서 Pretrained 모델의 accuracy 회복 (동일한 learning rate schedule 121epoch).
% 3. FC Layer를 Freeze하고 conv 압축 시작
% - 1,2,3번째 layer (i_c < 128) 제외하고 압축
% - Tile SVD는 64x64 tile
% - Lr schedule : 0.01부터 시작 50epoch 마다 1/2로 decay, i.e lr = init_lr * (0.5 ** (epoch // 50)) 
% - dist_step = 300

\begin{table*}[h]
\begin{center}
\caption{Comparison on various low-rank approximation schemes of VGG19 (using CIFAR-10 dataset). To focus on convolution layers only, fully-connected layers are compressed by 8$\times$ and trained by occasional regularization. Then, fully-connected layers are frozen and convolution layers are compressed (except small layers of $S<128$) by Tucker decomposition or tiled SVD.} 
\label{table:vgg}
\begin{tabular}{cccccc}
\Xhline{2\arrayrulewidth}
\multicolumn{1}{c}{Comp. Scheme}  & Parameter &  Weight Size & FLOPs & Accuracy(\%) \\
\Xhline{2\arrayrulewidth}
 \mr{1}{Pre-Trained} & - & \mr{1}{18.98M} & \mr{1}{647.87M} & \mr{1}{92.37} \\
\hline
\mr{4}{Tucker\\ Decomposition \\ (Typical\\Scheme)}  & $R_c$=0.6~~  & 9.14M (2.08$\times$) & 319.99M (2.02$\times$) & 91.97 \\ 
                                & $R_c$=0.5~~  & 6.71M (2.83$\times$) & 235.74M (2.75$\times$) & 91.79 \\
                                & $R_c$=0.45 & 5.49M (3.45$\times$) & 191.77M (3.38$\times$) & 91.36 \\
                                & $R_c$=0.4~~  & 4.61M (4.11$\times$) & 161.60M (4.01$\times$)& 91.11 \\
\hline
\mr{5}{Tiled\\SVD\\(Occasional\\Regularization,\\ $pN\!R$=300)}  & 64$\times$64 ($r$=16) & 9.49M (2.00$\times$) & 316.28M (2.04$\times$) & 92.42 \\
                    & 64$\times$64 ($r$=11) & 6.52M (2.91$\times$) & 214.25M (3.02$\times$) & 92.33 \\
                    & 64$\times$64 ($r$=10) & 5.93M (3.20$\times$) & 193.85M (3.34$\times$) & 92.23\\
                    & 64$\times$64 ($r$=9)~~ & 5.55M (3.41$\times$) & 173.44M (3.73$\times$) & 92.22 \\
                    & 64$\times$64 ($r$=8)~~ & 4.74M (4.00$\times$) & 153.04M (4.33$\times$) & 92.07 \\
\Xhline{2\arrayrulewidth}                                                
\end{tabular}
\end{center}
\end{table*}

We compare Tucker decomposition trained by a typical fine-tuning process and tiled SVD trained by occasional regularization using the VGG19 model\footnote{https://github.com/chengyangfu/pytorch-vgg-cifar10} with CIFAR-10.
Since this work mainly discusses compression on convolution layers, fully-connected layers of VGG19 are compressed and fixed before compression of convolution layers (refer to Appendix for details on the structure of VGG19).
Except for small layers with $S<128$ (that presents small compression ratio as well), all convolution layers are compressed with the same compression ratio.
During 300 epochs to train convolution layers, learning rate is initially 0.01 and is then halved every 50 epochs.
In the case of tiled SVD, $pN\!R$ is 300 and tile size is fixed to be 64$\times$64 (recall that the choice of $pN\!R$ and tile size do not affect model accuracy significantly).
As described in Table~\ref{table:vgg}, while Tucker decomposition with conventional fine-tuning shows degraded model accuracy through various $R_c$, occasional-regularization-assisted tiled SVD presents noticeably higher model accuracy.

% Graph : ResNet34 + ImageNet에 대한 Tucker vs TiledSVD Test Accuracy 그래프 (3.2배 압축 상황)
% ResNet34 Structure
%o_c    i_c k k     # of layer  압축대상 (98.6%)
%64	    3   7 7     1           x
%64	    64  3 3     6           x
%128	64  3 3     1           x
%128	128 3 3     7           o
%256	128 3 3     1           o
%256	256 3 3     11          o
%512	256 3 3     1           o
%512	512 3 3     5           o
% learning recipe : 0.01 20epoch 0.001 30epoch 0.0001 30epoch (총 80epoch)
% Batch size = 128, 1epoch = 10010 step

% Tucker : 압축 대상 Layer의 S, T에 x 0.46(R_c) ==> 3.198x 압축 (전체 3.097x)
% Tile SVD
% - 64x64 tiled svd, r=10 ==> 3.2배 압축 (전체 3.104x)
% - S_D = 2000 step (더 실험 예정)

\begin{figure*}[h]
\begin{center}
    \centering
    \includegraphics[width=0.8\linewidth]{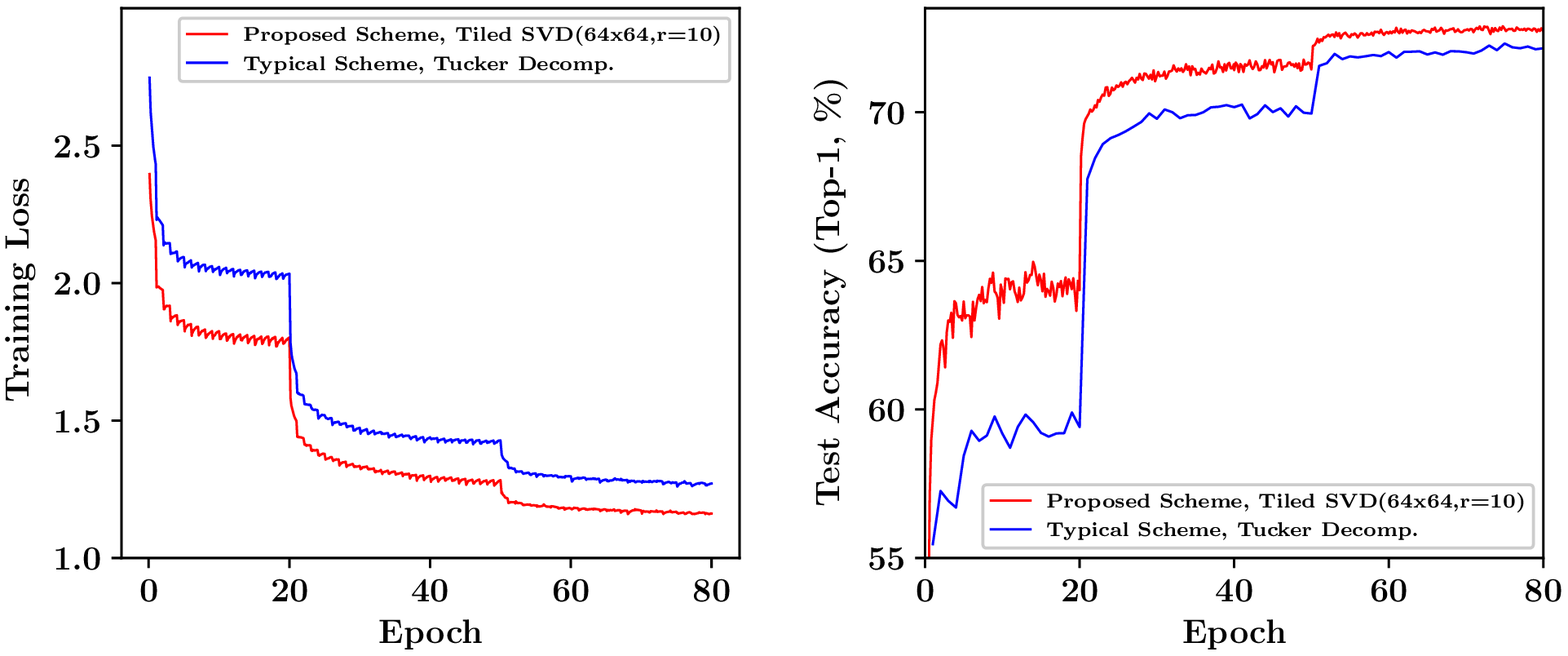}
\caption{Comparison of two compression schemes on training loss and (top-1) test accuracy of ResNet-34 model using ImageNet. $pN\!R$=500.} 
\label{fig:imagenet}
\end{center}
\end{figure*}

We also test our proposed low-rank approximation training technique with the ResNet-34 model\footnote{https://pytorch.org/docs/stable/torchvision/models.html} \cite{resnet} using the ImageNet dataset \cite{imagenet}.
A pre-trained ResNet-34 is fine-tuned for Tucker decomposition (with conventional training) or tiled SVD (with occasional regularization) using the learning rate of 0.01 for the first 20 epochs, 0.001 for the next 30 epochs, and 0.0001 for the remaining 30 epochs.
Similar to our previous experiments, the same compression ratio is applied to all layers except the layers with $S<128$ (such exceptional layers consist of 1.4\% of the entire model).
In the case of Tucker decomposition, selected convolution layers are compressed with $R_c=0.46$ to achieve an overall compression of $3.1\times$.
For tiled SVD, lowered matrices are tiled and each tile of (64$\times$64) size is decomposed with $r$=10 to match an overall compression of $3.1\times$.
As shown in Figure~\ref{fig:imagenet}, occasional-regularization-based tiled SVD yields better training loss and test accuracy compared to Tucker decomposition with typical training.
At the end of the training epoch in Figure~\ref{fig:imagenet}, tiled SVD and Tucker decomposition achieves 73.00\% and 72.31\% for top-1 test accuracy, and 91.12\% and 90.73\% for top-5 test accuracy, while the pre-trained model shows 73.26\% (top-1) and 91.24\% (top-5).

\section{Lowering Technique for CNNs}

\begin{figure*}[h]
\begin{center}
    \centering
    \includegraphics[width=1.0\linewidth]{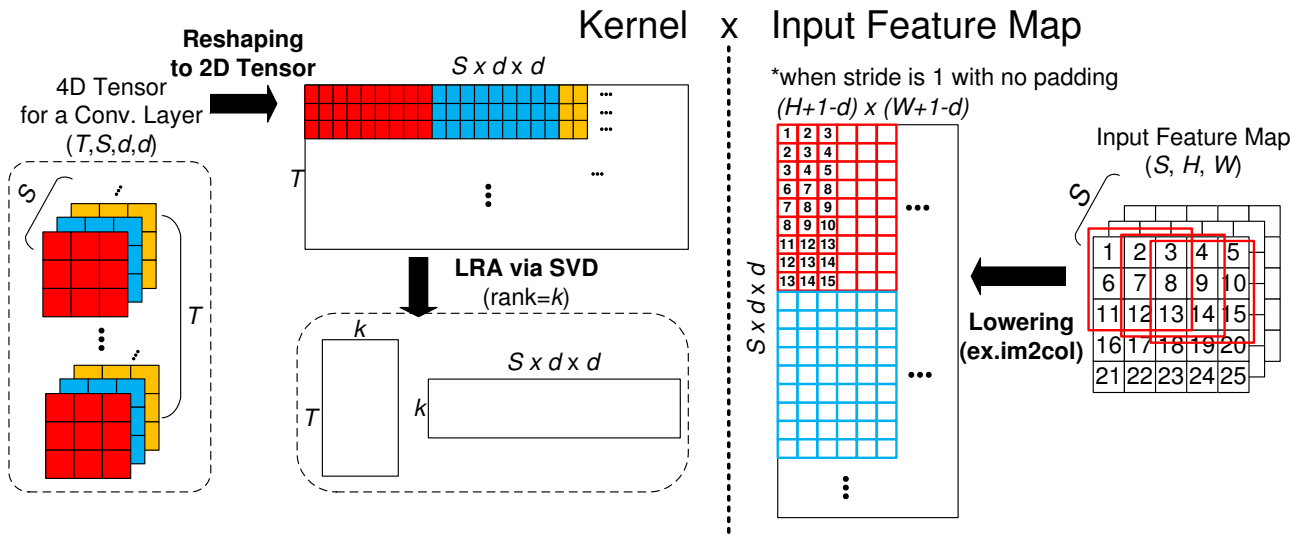}
\caption{An example of lowering technique using im2col.} 
\label{fig:lowering}
\end{center}
\end{figure*}

Figure~\ref{fig:lowering} describes a kernel matrix reshaped from a 4D kernel tensor and an input feature map matrix in the form of a Toeplitz matrix.
At the cost of redundant memory usage to create a Toeplitz matrix, lowering enables matrix multiplication which can be efficiently implemented by BLAS libraries.
A kernal matrix can be decomposed by 2D SVD.

\section{Model Descriptions for Low-Rank Approximation Experiments}

In this section, we describe model structures and layers selected for low-rank approximation experiments.
%PTB에 대한 내용도 추가해야함.
Small layers close to the input are not compressed because both weight size and compression rate are too small.

\begin{table*}[h]\begin{center}
\caption{Convolution Layers of ResNet-32 for CIFAR-10} 
\label{table:app_resnet32}
\begin{tabular}{crrrr|c}
\Xhline{2\arrayrulewidth}
\# of layers & $T$ & $S$ & $d$ & Weight Size & Decomposed \\
\hline
1	& 16	& 3		& 3	& 0.4K (~~0.1\%) & No \\
10	& 16	& 16	& 3	& 22.5K (~~5.0\%) & Yes \\
1	& 32	& 16	& 3	& 4.5K (~~1.0\%) & Yes \\
9	& 32	& 32	& 3	& 81.0K (18.0\%) & Yes \\
1	& 64	& 32	& 3	& 18.0K (~~4.0\%) & Yes \\
9	& 64	& 64	& 3	& 324.0K (71.9\%) & Yes \\
\hline
\multicolumn{4}{c}{\textbf{Total}} & 450.4K (100.0\%)  \\
\Xhline{2\arrayrulewidth}                                                
\end{tabular}\end{center}\end{table*}

\begin{table*}[h]\begin{center}
\caption{Convolution and Fully-connected (FC) Layers of VGG-19 for CIFAR-10} 
\label{table:app_vgg19}
\begin{tabular}{c|crrrr|c}
\Xhline{2\arrayrulewidth}
 Type & \# of layers & $T$ & $S$ & $d$ & Weight Size & Decomposed \\
\hline

\mr{8}{Conv.} & 1 &	64 &	3 & 3 &	0.002M (~~0.01\%) & No \\
 & 1 &	64 &	64 & 3 & 0.035M (~~0.18\%) & No \\ 
 & 1 &	128 &	64 & 3 & 0.070M (~~0.36\%) & No \\
 & 1 &	128 &	128 & 3 & 0.141M (~~0.72\%) & Yes \\
 & 1 &	256 &	128 & 3 & 0.281M (~~1.44\%) & Yes \\
 & 3 &	256 &	256 & 3 & 1.688M (~~8.61\%) & Yes \\
 & 1 &	512 &	256 & 3 & 1.125M (~~5.74\%) & Yes \\
 & 7 &	512 &	512 & 3 & 15.75M (80.37\%) & Yes \\
\hline				
%\mr{2}{FC} & \# of layers & input & output & & Weight Size & \mr{3}{Pre-trained \\ and Fixed} \\
%\cline{2-6}
\mr{2}{FC} & 2 & 512 & 512 & - & 0.500M (~~2.55\%) & \mr{2}{Yes \\ Yes} \\
& 1 & 512 & 10 & - & 0.005M (~~0.02\%) & \\
\hline			
\textbf{Total} & & & & & 19.597M (100.0\%)  \\
\Xhline{2\arrayrulewidth}                                                
\end{tabular}\end{center}\end{table*}

\begin{table*}[h]\begin{center}
\caption{Convolution Layers of ResNet-34 for ImageNet} 
\label{table:app_resnet34}
\begin{tabular}{crrrr|c}
\Xhline{2\arrayrulewidth}
\# of layers & $T$ & $S$ & $d$ & Weight Size & Decomposed\\
\hline
1 & 64 & 3 & 7 & 0.01M (~~0.04\%) & No \\
6 & 64 & 64 & 3 & 0.21M (~~1.05\%) & No \\ 
1 & 128	& 64 & 3 & 0.07M (~~0.35\%) & No \\ 
7 & 128	& 128 & 3 & 0.98M (~~4.90\%) & Yes \\
1 & 256	& 128 & 3 & 0.28M (~~1.40\%) & Yes  \\
11 & 256 & 256 & 3 & 6.18M (30.77\%) & Yes\\
1 & 512	& 256 & 3 & 1.13M (~~5.59\%) & Yes \\
5 & 512	& 512 & 3 & 11.25M (55.94\%) & Yes\\
\hline
Total & & & & 20.11M (100.0\%) & \\
\Xhline{2\arrayrulewidth}                                                
\end{tabular}\end{center}\end{table*}

\end{document}